\documentclass[journal]{IEEEtran}
\usepackage{amsfonts}
\usepackage{amsmath,graphicx,setspace,algorithm,algorithmic,multirow,cases,amssymb,subfigure}
\usepackage{lettrine}
\usepackage{multirow}
\usepackage{xcolor}
\usepackage{times}
\usepackage{graphicx}
\usepackage{amsmath}
\usepackage{indentfirst}
\usepackage{booktabs}
\usepackage{blindtext}
\usepackage{amsfonts,amssymb}
\usepackage{array}
\usepackage{tabularx,booktabs}
\usepackage{pgfplots}
\pgfplotsset{compat=1.18} 
\usepackage{subcaption}
\usepackage{tikz}
\usetikzlibrary{positioning}

\usepackage[
            pdfstartview=FitH,
            CJKbookmarks=true,
            bookmarksnumbered=true,
            bookmarksopen=true,
            linkcolor=red,
            anchorcolor=red,
            citecolor=green
            ]{hyperref}

\begin{document}
\title{Spectral Colorimetry Based on GMoE for Temperature Measurement in Explosion Fields}

\author{\IEEEauthorblockN{Jie Fang, Wenhao Li, Jianning Yang, and Nan Wang}
	
	\thanks{		
		J. Fang, W. Li, and J. Yang are with the School of Telecommunication and Information Engineering, Xi'an University of Posts and Telecommunications, Xi'an 710121, Shaanxi, P. R. China. J. Fang is also the Chief Scientist of Hangzhou Halome Cloud Technology Company Limited, Hangzhou 310020, Zhejiang, P. R. China. N. Wang is with the School of Information Science and Technology, Hainan Normal University, Haikou 571158, China.}}
\maketitle

\begin{abstract}
Spectroscopic thermometry in explosion fields has attracted considerable attention in related engineering domains, primarily because it can mitigate the interference of traditional contact methods with temperature distributions and address the limitations of high-temperature measurement scenarios. However, existing physics-model-driven methods typically rely on regressing a single radiance-ratio pair, which limits their adaptability to wide-range temperature estimation. This limitation arises from notable variations in the mapping between grayscale ratios and radiance ratios at different temperatures. In contrast, existing data-driven methods often suffer from limited performance due to the scarcity of training data, a challenge exacerbated by the sensitivity of explosion fields. In this work, we propose a gated mixture of experts (GMoE) framework that integrates a divide-and-conquer strategy in deep learning with physical prior knowledge for temperature estimation, thereby decomposing the complex task into multiple expert-specific subtasks. Specifically, each expert is developed using physically calibrated polynomial functions to model the relationship between grayscale ratios and radiance ratios for a given pair of spectral bands, with temperature then estimated using a physical model incorporating the corresponding parameters. Additionally, a gating network is designed to adaptively weight the outputs of different experts and generate the final temperature estimate. Experimental results demonstrate that the proposed framework achieves the best performance when implemented with the Transformer-based gating network, yielding a mean RMSE of $12.70 \pm 1.32$ K and a mean MAE of $7.96 \pm 0.86$ K. Compared with conventional spectral colorimetry, the proposed framework reduces the mean RMSE and MAE by 63.4\% and 70.8\%, respectively. The source code and dataset are available at our GitHub repository: \url{https://github.com/123-lwh/gmoe-spectral-colorimetry-dataset}.
\end{abstract}

\IEEEpeerreviewmaketitle

\section{Introduction}\label{introduction}
\begin{figure}[t]
\centering
    \includegraphics[width=1.00\linewidth]{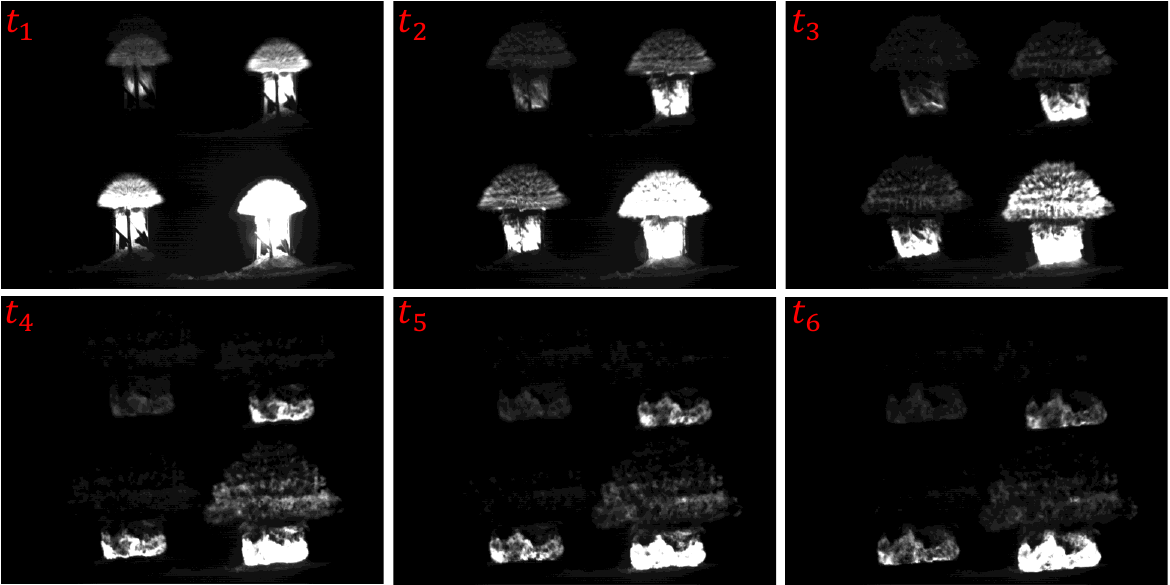}
\caption{Spectral variation process of the explosion fields.}
\label{motivation}
\end{figure}

Temperature is a key parameter of the explosion field \cite{girard2014multiwavelength}. Measuring the explosion-field temperature enables not only the reconstruction and analysis of explosion-process evolution and efficiency but also the investigation of its damage effects and explosion mechanisms \cite{vuelban2015radiometric}. Conventional temperature estimation methods in explosion fields primarily rely on infrared (IR) imaging techniques, thermocouples, and pyrometers. These methods are typically grounded in the blackbody radiation approximation or single-spectral-band radiance measurements to directly estimate the temperature value via established physical laws. However, such approaches exhibit inherent limitations. First, they are susceptible to environmental noise, background radiation, and variations in measurement angle. Second, stringent instrument calibration is indispensable, especially under high-temperature, dynamic, or complex operational scenarios, while the calibration process is time-consuming and difficult to maintain with high accuracy. Third, conventional methods often struggle to model non-ideal materials or temperature-dependent radiance, leading to deviations between the estimated and true temperature values. For example, in high-performance ceramics or metallic materials, radiance exhibits significant variations with temperature and wavelength, which poses notable challenges for physics-based temperature inversion. As illustrated in Fig. \ref{motivation}, the multispectral images of the explosion field exhibit evident spatial and temporal intensity variations, reflecting the dynamic and wavelength-dependent characteristics of explosion-field radiation.

With the rapid advancement of multispectral imaging technologies, acquiring radiation data across multiple spectral bands, ranging from visible to mid- and far-infrared (MIR-FIR) \cite{daffara2011wide}, offers a novel pathway for temperature estimation. The recorded image intensity in each spectral band is governed by the thermal radiation emitted by the target, which is determined by its temperature according to thermal radiation laws \cite{usamentiaga2014infrared}. Thus, multispectral images provide wavelength-dependent radiative information for temperature estimation. Compared with single- or dual-spectral-band measurements, multispectral temperature estimation can leverage a broader spectral range, thereby enabling more comprehensive characterization of spectral radiance variations across wavelengths and temperatures. Such richer spectral information not only enhances temperature estimation accuracy but also mitigates the impact of environmental noise and varying measurement conditions. Nevertheless, multispectral temperature estimation remains challenging because real-material radiance is nonlinear and temperature-dependent, while practical measurements are affected by noise, background radiation, instrument response, band selection, and radiative coupling effects.

Against this backdrop, data-driven approaches have emerged as a growing research focus. Early studies primarily adopted traditional regression models, such as polynomial fitting, support vector regression (SVR) \cite{hearst1998support}, and locally weighted regression (LWR) \cite{cleveland1988locally}, to establish a quantitative nonlinear mapping between multispectral radiance data and temperature. While these methods have achieved certain success under laboratory conditions, their generalization performance remains limited in the presence of heterogeneous materials, dynamic measurement scenarios, or environments with high measurement noise. To address these limitations, deep neural networks (DNNs) have been increasingly leveraged for temperature estimation \cite{xing2023graphical,mao2025neural}, as they can automatically learn complex nonlinear features through end-to-end training, thus improving prediction accuracy. However, the acquisition of real-world explosion-field data is cost-prohibitive, and the resulting data scarcity can readily lead to overfitting, thereby limiting the performance of deep learning models.

In addition, multispectral thermometry exhibits temperature-dependent radiance characteristics across bands: owing to different spectral responses and sensitivity ranges, the informativeness and noise tolerance of each wavelength band (or band ratio) may vary with the temperature regime. Consequently, a single physical inversion with fixed assumptions often fails to account for non-ideal effects, while a purely data-driven single-model regressor is forced to fit heterogeneous radiance–temperature relationships over a wide range using one global function. This can lead to an average-fit compromise and suboptimal fitting in certain regimes, especially when data are limited or when band-specific degradation occurs. This observation motivates the proposed expert-based formulation, in which conventional wavelength-pair colorimetric thermometry models are constructed as individual experts, and the gating module adaptively combines their outputs according to the current multispectral observation.

To address these challenges, we utilize a high-precision standard blackbody radiator and a split-aperture multispectral imaging system to construct a dataset with a temperature range similar to that of explosion fields. The resulting dataset contains 165 samples with annotated temperature labels ranging from 1000$^\circ$C to 3000$^\circ$C. On this basis, this paper develops a physics-guided Gated Mixture of Experts (GMoE) framework for multispectral temperature estimation. This framework integrates physical constraints on spectral radiance with the adaptive learning capability of the Mixture of Experts (MoE), thereby enabling the expert models to capture diverse radiance patterns or spectral dependencies. Within this framework, the gating network dynamically assigns weights to different experts based on the observed multispectral data. By integrating domain knowledge with data-driven modeling, the proposed approach maintains physically consistent behavior while providing improved temperature prediction performance in the evaluated experiments. The main contributions of this work are summarized as follows:
\begin{itemize}
      \item We constructed a temperature estimation dataset using a high-precision standard blackbody radiator and a split-aperture multispectral imaging system, which contains 165 samples with temperatures ranging from 1000$^\circ$C to 3000$^\circ$C.
	\item We construct wavelength-pair colorimetric thermometry models as individual experts, where each expert provides a temperature estimate based on the calibrated mapping from grayscale ratios to radiance ratios.
       \item We develop a gating-based fusion strategy that adaptively combines the outputs of different experts according to the multispectral observation, thereby improving temperature estimation accuracy.
       \item We conduct a series of experiments to demonstrate that the proposed method outperforms the existing model-driven and data-driven methods, and further verify the rationality of our proposed hybrid collaborative temperature estimation framework.
\end{itemize}

The remainder of this paper is organized as follows. Section \ref{proposedmethod} details the proposed method, including preliminaries, data preprocessing, and the proposed framework. Subsequently, Section \ref{experiments} presents the experimental results and corresponding analysis. Finally, Section \ref{conclusion} concludes the paper and outlines future research directions.

\section{Related Works}\label{relatedworks}
\subsection{Temperature estimation methods based on physical model}
Accurate measurement of explosion-related parameters (e.g., temperature fields, gas properties) has evolved significantly over the past two decades. These advances in data processing, optical sensing, and spectroscopy have helped address challenges such as sparse data, transient dynamics, and harsh environments. The key contributions in this field are reviewed chronologically below.

\subsubsection{2007–2008: Foundational Advances in High-Speed Spectroscopy}
Early work focused on enhancing the speed and reliability of spectral measurements for gas analysis in combustion and explosion scenarios. Kranendonk et al. \cite{kranendonk2007high} pioneered a high-speed, low-noise molecular absorption spectroscopy system using a novel FDML laser-based time-multiplexed design, laying the groundwork for real-time gas property monitoring. The following year, Sanders et al. \cite{sanders2008frequency} built on this progress by developing and validating multiple laser sources for high-speed spectral measurement of gas properties in combustion flows, expanding the applicability of laser-based techniques to dynamic environments.

\subsubsection{2014–2017: Radiation Thermometry and Trace Detection Innovations}
The mid-2010s saw parallel advancements in optical temperature measurement and trace explosive detection. In 2014, Nie et al. \cite{nie2014temperature} introduced a radiation thermometry method for reconstructing two-dimensional temperature fields of premixed gas explosion flames from high-speed camera images, incorporating region-specific emissivity corrections and validating against thermocouple data to improve accuracy. Concurrently, prior studies have explored opto-calorimetric IR spectroscopy and laser absorption spectroscopy, combined with statistical analysis, for sensitive and selective detection of adsorbed explosives and gas species \cite{zandieh2016sensitive, wang2017tunable}.

\subsubsection{2016–2021: Data-Driven Reconstruction and In-Situ Calibration}
Alongside advances in optical techniques, data-driven methods emerged to address sparse sensor data limitations, while in-situ calibration solutions were developed for harsh conditions. Chen et al. \cite{chen2016overview} demonstrated in 2016 that orthogonal matching pursuit improves modeling efficiency over least-squares fitting, and Bayesian model fusion enables accurate spatial distribution reconstruction from sparse sensor data, thereby overcoming a key bottleneck in traditional measurement systems. In 2021, Tai et al. \cite{tai2021flame} advanced in-situ measurement by achieving accurate temperature field reconstruction through spectral emissivity calibration of intrinsic element particles using existing high-temperature equipment, eliminating the need for external calibration tools in extreme environments.

\subsubsection{2022–2023: Transient Temperature Measurement and Hardened Sensing}
Recent years have focused on optimizing techniques for transient explosion dynamics and ruggedized sensing. In 2022, Yao et al. \cite{yao2022explosion} employed two-color pyrometry to quantify transient temperature field dynamics of augmented emulsion explosives, emphasizing time-resolved behavior critical for explosion mechanism analysis. 2023 saw two notable contributions: Wang et al. \cite{wang2023development} designed a multispectral pyrometer with synchronous spectral-splitting technology to capture four-wavelength flame images simultaneously, integrating edge-based image registration for high spatiotemporal resolution temperature reconstruction; meanwhile, Schwartz et al. \cite{schwartz2023near} developed a hardened optical probe paired with laser absorption spectroscopy to quantify temperature in explosive fireballs, addressing mechanical and thermal robustness for real-world testing. LoCurto et al. \cite{locurto2023broadband} also applied hyperspectral water vapor absorption spectroscopy to laboratory-scale explosives in 2023, enabling detailed analysis of detonation-related water vapor dynamics.

\subsubsection{2024–2025: High-Speed Imaging and Extreme-Condition Adaptations}
The latest research (2024–2025) focuses on refining high-speed imaging and adapting methods to challenging environments. In 2024, Yu et al. \cite{yu2024novel} introduced a two-dimensional high-speed temperature measurement system, based on blackbody radiation theory and enhanced Bayer array interpolation to eliminate artifacts in transient explosion field spatiotemporal capture. Most recently, Pei et al. \cite{pei2025multi} developed a high-precision multispectral radiometric method tailored to wide-range explosion temperature measurement under harsh conditions (e.g., smoke, dust), addressing lingering environmental interference challenges.

\subsection{Temperature estimation methods based on data driven}
Beyond this chronological overview, recent studies and representative earlier works have further advanced techniques in temperature reconstruction, combustion simulation, and data analysis by leveraging both traditional spectroscopic methods and emerging machine learning (ML) and deep learning (DL) approaches to address unresolved challenges like noise interference, limited view angles, and computational efficiency.

\subsubsection{Traditional Spectroscopic and Thermometric Techniques: Foundational and Enhanced Methods}
Spectroscopic analysis has long been a cornerstone for flame temperature measurement, with early work establishing core principles. Broida et al. \cite{broida1957spectroscopic} pioneered the determination of flame electronic temperatures through spectroscopic analysis of iron emission spectra, laying the groundwork for optical temperature sensing in combustion systems. Decades later, Kun et al. \cite{kun2013development} enhanced multi-target multispectral thermometry by introducing a brightness temperature approximation method, mitigating the critical challenge of initial value selection in secondary measurement and improving the reliability of multi-target sensing. Meanwhile, advancements in multispectral and infrared thermometry have addressed specific scene needs. Pei et al. \cite {pei2025multispectral} employed four-wavelength multispectral thermometry integrated with blackbody calibration and Powell optimization, achieving accurate two-dimensional candle flame temperature reconstruction and demonstrating the value of optimization algorithms in refining multispectral results. For explosive fireball monitoring, Hobbs et al. \cite {hobbs2022high} showed that high-speed infrared radiation thermometers capture invaluable early-stage average surface temperatures over a wide field-of-view in confined blasts, with data corroborated by pressure gauges, thereby filling a gap in pre-quasi-static pressure onset measurement.

\subsubsection{Deep Learning and Machine Learning: Transforming Reconstruction and Simulation}
The rise of DL and ML has revolutionized combustion parameter analysis, enabling efficient and accurate inversion, simulation, and classification. Deng et al. \cite{deng2020deep} pioneered temperature field reconstruction via nonlinear tomographic absorption spectroscopy, utilizing deep belief and recurrent neural networks while providing systematic algorithmic comparisons, thereby setting a benchmark for DL-driven tomographic methods. In \cite{an2020deep}, An et al. developed a U-Net-based framework for turbulent-combustion simulation, and a related study used neural-network surrogates to accelerate inversion. Wei et al. \cite{wei2020deep} addressed limited-view inversion in mid-infrared methane absorption imaging, while Ren et al. \cite{ren2019machine} extended learning-based analysis to multi-parameter retrieval (temperature and species) and combustion-regime identification (GFRI-based CNN) for turbulent CH4/air flames. 

\subsubsection{Algorithm Optimization and Noise Suppression: Enhancing Measurement Robustness}
To improve the robustness of optical and data-driven measurements, researchers have focused on algorithmic refinements for noise reduction and parameter estimation. Dreyer et al. \cite{dreyer2019improved} developed an Abel inversion method that fits the line-of-sight projection of a predefined intensity distribution, significantly reducing experimental noise in axisymmetric flame projection analysis, which represents a key advancement for accurate flame profile reconstruction. For soot sensing, learning-based methods have been integrated with imaging/attenuation measurements to estimate soot propensity and soot volume fraction with improved practicality and accuracy \cite{pino2018soot, rodriguez2021retrieving}. More broadly, ML has also been applied to combustion-video analytics and combustion-state prediction, including unsupervised clustering for high-speed dynamics characterization, CNN-based mode classification from raw pressure signals, and hybrid deep models for complex prediction tasks \cite{debus2020high, zhu2019convolutional, han2021prediction}.

\subsection{Mixture of experts}
The Mixture-of-Experts (MoE) model has emerged as a pivotal paradigm for scaling neural networks efficiently, leveraging sparse activation to balance computational cost and performance. However, challenges such as load imbalance, representation collapse, inefficient training and inference, and limited multimodal adaptability have motivated extensive research on architectural refinements, optimization strategies, and application-specific adaptations. Recent works have advanced both foundational designs and practical deployment.

\begin{figure*}[t]
\centering
    \includegraphics[width=1.00\linewidth]{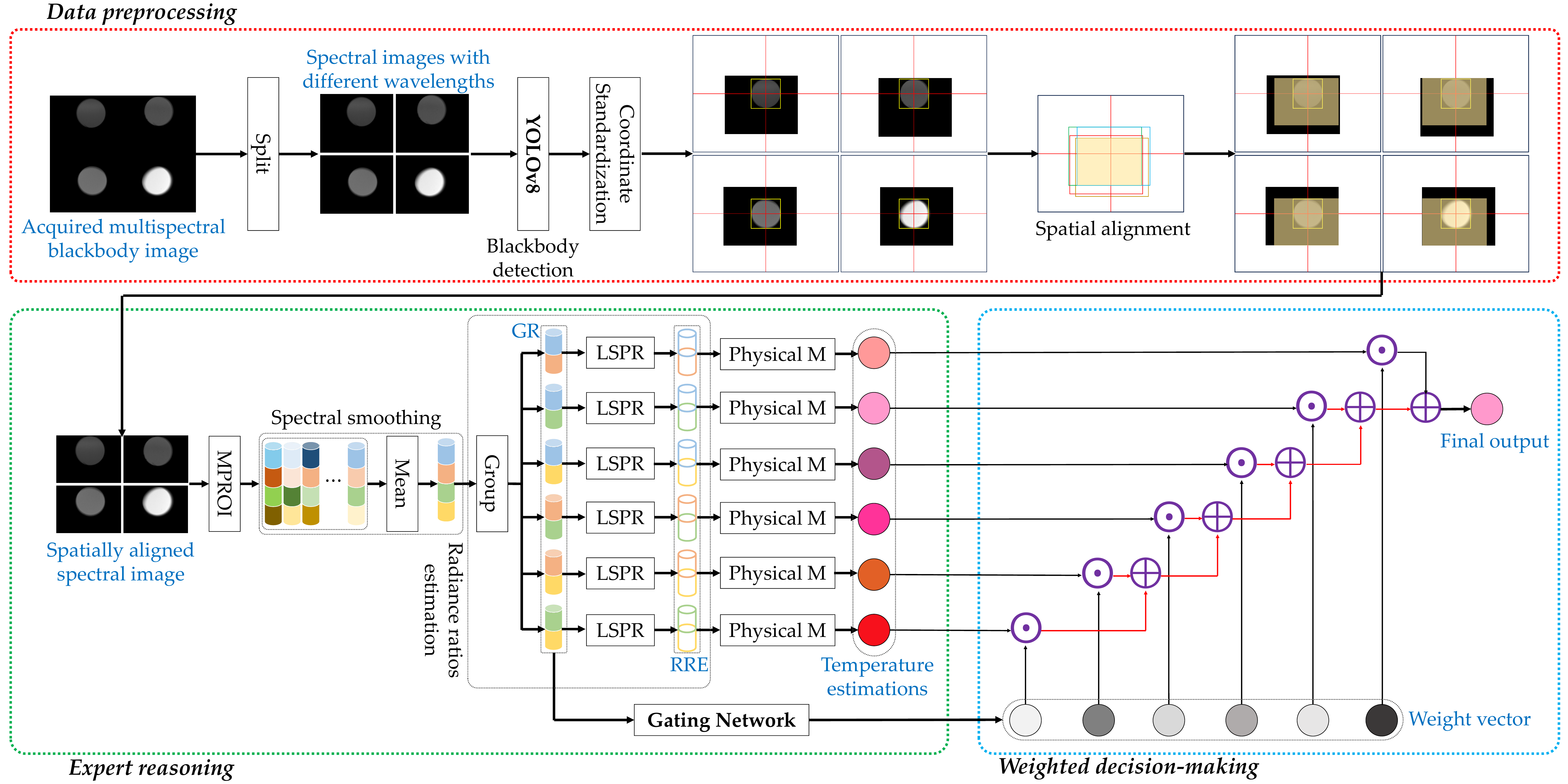}
\caption{The processing procedure of the proposed method, where MPROI denotes the multi-point region of interest samping. LSPR denotes the least squares polynomial regression. GR denotes the grayscale ratio. RRE denotes the radiance ratio estimation. Physical M denotes the physical model.}
\label{flowchart}
\end{figure*}

Recent MoE research spans diverse directions, mainly focusing on architectural design, routing refinement, expert specialization, and training stability, such as mitigating representation collapse. Representative studies include several vision-oriented MoE designs \cite{enzweiler2011multilevel, riquelme2021scaling, chen2022towards, chi2022representation, gale2023megablocks}. In parallel, efficiency-oriented optimizations have been developed to alleviate load imbalance and improve training or inference efficiency \cite{zhou2022mixture, wu2024multi, hwang2023tutel, rajbhandari2022deepspeed}. MoE has also been extended to multimodal and deployment settings through robust fusion mechanisms for missing or noisy modalities, as well as parameter- and compute-efficient expert formulations \cite{mustafa2022multimodal, han2024fusemoe, wang2025decoupled, huang2024toward, huang2025ders}.

Part A motivates our use of physics-based colorimetric inversion as the core estimators, while also indicating that the reliability of different wavelength pairs can vary across temperature ranges and measurement conditions. Part B further suggests that purely data-driven mappings may be sensitive to domain shifts and harsh disturbances, motivating an adaptive mechanism that improves robustness without discarding physical interpretability. Part C, in turn, provides the MoE principle for conditional model combination. Accordingly, we formulate a compact MoE in which each expert corresponds to a physics-based estimator for one spectral-pair input, and a gating network adaptively fuses these estimates to produce the final temperature.

\section{Proposed Method}\label{proposedmethod}

\subsection{Overview}
In this study, we propose a hybrid expert-based framework that integrates a physics-informed prior model with a data-driven component to achieve high-precision temperature estimation in explosion fields under complex scenarios and limited training data. Instead of learning a single global mapping from multispectral observations to temperature, the proposed framework constructs conventional wavelength-pair colorimetric thermometry models as individual experts and uses a compact gating network to adaptively combine their outputs. In this design, the pre-fitted physical experts provide temperature estimates with clear colorimetric interpretation, while the trainable gating network adjusts the contribution of each expert according to the current multispectral observation. The overall workflow is illustrated in Fig.~\ref{flowchart} and consists of data preprocessing, expert reasoning, and weighted decision-making. Specifically, in the {\bf{data preprocessing stage}}, YOLOv8 is used to accurately localize each spectral channel in explosion-field images, and a multipoint ROI sampling strategy is applied to yield spatially consistent, low-noise grayscale ratios, providing a reliable foundation for subsequent temperature estimation. In the {\bf{expert reasoning stage}}, polynomial regression via least squares is used to fit the functional mapping from grayscale ratios to radiance ratios for each pair of spectral bands. The resulting polynomial experts are pre-fitted and kept fixed during training. The explosion-field-specific colorimetric thermometry model then derives temperature estimates by incorporating the corresponding polynomial parameters. In the {\bf{weighted decision-making stage}}, a gating network is adopted to adaptively weight each expert’s temperature estimate and fuse them into a more accurate final result. Overall, this hybrid design combines fixed physically interpretable expert modeling with a compact trainable gating module, thereby reducing the dependence on large-scale labeled data.

\subsection{Data preprocessing}
\subsubsection{Image split}
The data used in this work were acquired with a split-aperture multispectral imaging system, where spectral bands are distributed in a $2\times2$ grid on a single image. Accordingly, we split the acquired image into $2\times2$ blocks to obtain spectral images with different wavelengths, as defined in Equation (\ref{split}).

\begin{equation}
\label{split}
\begin{array}{c}
\begin{aligned}
{\bf{I}} &= \left( {\begin{array}{*{20}{c}}
{{{\bf{I}}_1}}&{{{\bf{I}}_2}}\\
{{{\bf{I}}_3}}&{{{\bf{I}}_4}}
\end{array}} \right)\\
 &= \left( {\begin{array}{*{20}{c}}
{{{\bf{I}}_{\left( {1:\left\lfloor {\frac{H}{2}} \right\rfloor ,1:\left\lfloor {\frac{W}{2}} \right\rfloor } \right)}}}&{{{\bf{I}}_{\left( {1:\left\lfloor {\frac{H}{2}} \right\rfloor ,\left\lfloor {\frac{W}{2}} \right\rfloor  + 1:W} \right)}}}\\
{{{\bf{I}}_{\left( {\left\lfloor {\frac{H}{2}} \right\rfloor  + 1:H,1:\left\lfloor {\frac{W}{2}} \right\rfloor } \right)}}}&{{{\bf{I}}_{\left( {\left\lfloor {\frac{H}{2}} \right\rfloor  + 1:H,\left\lfloor {\frac{W}{2}} \right\rfloor  + 1:W} \right)}}}
\end{array}} \right)
\end{aligned}
\end{array}
\end{equation}
where $\bf{I}$ denotes the original acquired image. ${\bf{I}}_1\), \({\bf{I}}_2$, ${\bf{I}}_3$, and ${\bf{I}}_4$ denote the spectral images distributed in the top-left, top-right, bottom-left, and bottom-right regions, respectively. $W$ and $H$ denote the width and height of the original acquired image, respectively. $\left\lfloor  \cdot  \right\rfloor$ denotes the floor function.

\subsubsection{Blackbody detection}
Even though the split spectral bands should be strictly spatially aligned, they often exhibit deviations in practice due to the alignment and assembly errors of the imaging system. To achieve pixel-level spatial alignment, it is necessary to calibrate the multispectral images for spatial consistency using a reference object. Therefore, the round blackbody, especially its center, can be used as a natural reference object. We accurately detect the blackbody centers in different spectral images using YOLOv8 \cite{yaseen2408yolov8} to establish a reference location, and the detection process is defined as Equation (\ref{yolov8}).

\begin{equation}
\label{yolov8}
\left( {\begin{array}{*{20}{c}}
{x_i^c}\\
{y_i^c}
\end{array}} \right) = {\rm{YOLOv8}}\left( {{{\bf{I}}_i}} \right)
\end{equation}
where $\left( {\begin{array}{*{20}{c}}
{x_i^c}\\
{y_i^c}
\end{array}} \right)$ denotes the center of the blackbody in the $i_{th}$ spectral band. In addition, YOLOv8 outputs a bounding box with width $W_i$ and height $H_i$, from which an effective radius $R_i$ is defined as $R_i=\min\!\left(W_i, H_i\right)/2$. The detected center  $\left( {\begin{array}{*{20}{c}}
		{x_i^c}\\
		{y_i^c}
\end{array}} \right)$ is used as the anchor for subsequent cross-band spatial alignment of the four spectral channels, and $R_i$ is used to constrain the sampling region within the blackbody area to suppress background interference during grayscale extraction.

\subsubsection{Coordinate standardization}

After obtaining the reference points in different spectral bands, we use Equation (\ref{uov}) to transform the original four coordinate systems a unified coordinate system.

\begin{equation}
\label{uov}
\left( {\begin{array}{*{20}{c}}
{{u_i}}\\
{{v_i}}
\end{array}} \right) = \left( {\begin{array}{*{20}{c}}
{{x_i}}\\
{{y_i}}
\end{array}} \right) - \left( {\begin{array}{*{20}{c}}
{x_i^c}\\
{y_i^c}
\end{array}} \right)
\end{equation}
where $\left( {\begin{array}{*{20}{c}}
{{x_i}}\\
{{y_i}}
\end{array}} \right)$ denotes the coordinate in the $i_{th}$ original coordinate system, whereas $\left( {\begin{array}{*{20}{c}}
{{u_i}}\\
{{v_i}}
\end{array}} \right)$ denotes the corresponding coordinate in the common coordinate system.

\subsubsection{Spatial alignment}
In the common coordinate system, spectral images from different bands may lose shared information due to positional deviations. Such band-to-band misregistration is particularly critical for ratio-based thermometry, as misalignment may corrupt intensity ratios and lead to erroneous temperature estimation \cite{vallabh2022melt}. To address this, we use Equation (\ref{iou}) to extract the spatial regions with complete spectral information, which serve as valid data for subsequent processing.

\begin{equation}
\label{iou}
{{\bf{I}}_{{\mathop{\rm IoU}\nolimits} }} = {{\bf{{I}}}_{uov}}\left( {{u_t}:{u_b},{v_l}:{v_r},:} \right)
\end{equation}
where ${{\bf{I}}_{{\mathop{\rm IoU}\nolimits} }}$ denotes the effective region in the common $uov$ coordinate system. ${{\bf{{I}}}_{uov}}$ denotes the stacked spectral image aligned according to the reference point. $\left( {\begin{array}{*{20}{c}}
{{u_t}}\\
{{v_l}}
\end{array}} \right)$ and $\left( {\begin{array}{*{20}{c}}
{{u_b}}\\
{{v_r}}
\end{array}} \right)$ denote the top-left and bottom-right points of the effective region in $uov$ coordinate system, respectively, and they are defined as Equation (\ref{uovedge}).

\begin{equation}
\label{uovedge}
\left( {\begin{array}{*{20}{c}}
{{u_t}}\\
{{u_b}}\\
{{v_l}}\\
{{v_r}}
\end{array}} \right) = \left( {\begin{array}{*{20}{c}}
{\max \left( { - x_1^c, - x_2^c, - x_3^c, - x_4^c} \right)}\\
{\min \left( {H - x_1^c,H - x_2^c,H - x_3^c,H - x_4^c} \right)}\\
{\max \left( { - y_1^c, - y_2^c, - y_3^c, - y_4^c} \right)}\\
{\min \left( {W - y_1^c,W - y_2^c,W - y_3^c,W - y_4^c} \right)}
\end{array}} \right)
\end{equation}

\subsection{Experts' reasoning}
\subsubsection{MPROI}

\begin{figure}[htbp]
\centering
    \includegraphics[width=1.00\linewidth]{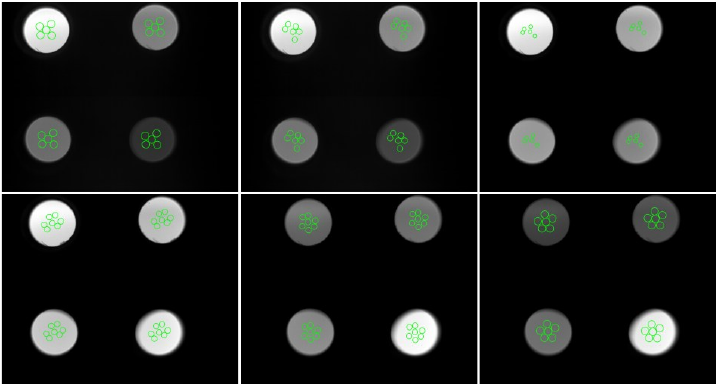}
\caption{Examples of MPROI sampling generated by the preprocessing pipeline. 
	Green circles denote the sampled ROIs whose centers are constrained within a neighborhood of radius $R$, while the ROI radius is randomly chosen from $\{10,15,20\}$ pixels with $K\in\{5,6,7\}$ per image.}
\label{mproi1}
\end{figure}

To alleviate the impact of spatially non-uniform brightness within each channel, we adopt a multi-point region-of-interest (MPROI) sampling strategy as a preprocessing step prior to polynomial fitting. Using the YOLOv8-detected channel center as the base point, $K$ circular regions of interest are randomly sampled within its adjacent area with radius $R$, and the strategy is defined in Equation (\ref{mproi}).

\begin{equation}
\label{mproi}
{\left\{ {{{\bf{s}}^c}} \right\}^K} = \mathop  \cup \limits_{k = 1}^K {{\bf{I}}_{uov}}\left( {{\pi _k}\left( {2{R^u}} \right) - {R^u},{\pi _k}\left( {2{R^v}} \right) - {R^v}} \right)
\end{equation}

where ${\left\{ {{{\bf{s}}^c}} \right\}^K}$ denotes the collection of selected circular regions with scale $K$. $\pi \left( x \right)$ is defined as a random integer generation function that produces integers within the interval $[0, x]$. ${{R^u}}$ and ${{R^v}}$ denote the neighborhood radius in two directions. In our implementation, we randomly set $K\in\{5,6,7\}$ per image, and the radius of each circular ROI is randomly chosen from $\{10,15,20\}$ pixels. For each ROI, the mean grayscale intensity is computed per channel, and wavelength-pair grayscale ratios are then formed from the ROI-wise means for subsequent polynomial fitting. Candidate polynomials are evaluated on a validation set, and the one achieving the minimum validation error is selected. Compared with single-ROI sampling, MPROI yields multiple local averages within the valid neighborhood, which reduces the variance of ratio estimation and mitigates occasional local outliers, thereby improving the stability of polynomial fitting and selection. Examples are shown in Fig. \ref{mproi1}.

\noindent where the regions outlined in green correspond to the selected regions.

\subsubsection{Spectral smoothing}
To obtain a representative spectral vector for each blackbody spectral image, we compute the average spectral vector over all selected pixels, which is defined as Equation (\ref{smooth}).

\begin{equation}
\label{smooth}
{{\bf{s}}_{\rm{m}}} = \frac{1}{K}\sum\limits_{k = 1}^K {\sum\limits_{\left( {i,j} \right) \in \mathcal N\left( k \right)} {{{\bf{s}}^c_k}\left( {i,j} \right)} }
\end{equation}
where ${\mathcal N\left( k \right)}$ denotes the coordinate set of $k_{th}$ selected circular region.

\subsubsection{Radiance ratios estimation}

Due to the nonlinear relationship between grayscale ratios and their corresponding radiance ratios, least-squares polynomial regression is adopted to estimate radiance ratios, which is defined as Equation (\ref{reijhat}).
\begin{equation}
\label{reijhat}
{\hat r_{ij}^r = \hbar \left( {r_{ij}^g\left| {\dot \Theta _{ij}^\hbar } \right.} \right)}
\end{equation}
where $r_{ij}^g$ and $\hat r_{ij}^r$ denote the grascale ratio and estimated radiance ratio of the $i_{th}$ and $j_{th}$ spectral bands, respectively. $\hbar \left(  \cdot  \right)$ denotes the polynomial function. $r_{ij}^g$ is defined as Equation (\ref{rijg}).

\begin{equation}
\label{rijg}
r_{ij}^g = \frac{{{{\bf{s}}_{\rm{m}}}\left( i \right)}}{{{{\bf{s}}_{\rm{m}}}\left( j \right)}}
\end{equation}
$\Theta _{ij}^\hbar$ in Equation (\ref{reijhat}) denotes the optimal parameters of corresponding polynomial function, which can be optimized with Equation (\ref{thetah}).

\begin{equation}
\label{thetah}
\dot \Theta _{ij}^\hbar  = \arg \mathop {\min }\limits_{\Theta _{ij}^\hbar } \frac{1}{N}{\left( {\hbar \left( {r_{ij}^g\left( n \right)\left| {\Theta _{ij}^\hbar } \right.} \right) - r_{ij}^e\left( n \right)} \right)^2}
\end{equation}
where $N$ denotes the number of samples in the training set composed of spectral images and their corresponding annotated temperatures. 
$r_{ij}^e$ denotes the theoretical value of radiance ratio, where Planck's blackbody radiation law under the Wien approximation is adopted \cite{zhao2025}, and $r_{ij}^{e}$ is defined as Equation (\ref{reijtrue}).

\begin{equation}
\label{reijtrue}
r_{ij}^e = \frac{{\varepsilon \left( {{\lambda _i}} \right) \cdot \frac{{{c_1}}}{{\lambda _i^5}} \cdot {e^{ - \frac{{{c_2}}}{{{\lambda _i}T}}}}}}{{\varepsilon \left( {{\lambda _j}} \right) \cdot \frac{{{c_1}}}{{\lambda _j^5}} \cdot {e^{ - \frac{{{c_2}}}{{{\lambda _j}T}}}}}}
\end{equation}
where $\lambda_i$ and $\lambda_j$ respectively denote the wavelengths corresponding to the $i_{\text{th}}$ and $j_{\text{th}}$ spectral bands. $c_1$ is defined as the first radiation constant, equal to $1.19104282 \times 10^{-16}\ \text{W}\cdot\text{m}^{-2}\cdot\text{Sr}^{-1}$. $c_2$ is the second radiation constant, with a value of $1.4387752 \times 10^{-2}\ \text{m}\cdot\text{K}$. $T$ stands for the annotated temperature value, and $\varepsilon(\cdot)$ denotes the spectral emissivity at wavelength $\lambda$ (for brevity written as $\varepsilon(\lambda)$, while in general it may depend on both wavelength and temperature, i.e., $\varepsilon=\varepsilon(\lambda,T)$). Solving Equation (\ref{reijtrue}) with respect to $T$ yields the temperature inversion with an explicit emissivity-ratio term, as given in Equation (\ref{tempij_full}):
\begin{equation}
	\label{tempij_full}
	{{\hat T}_{ij}} =
	\frac{{{c_2}\cdot\left(\frac{1}{{\lambda_j}}-\frac{1}{{\lambda_i}}\right)}}
	{{\ln\left(\hat r_{ij}^e\right)+5\ln\left(\frac{\lambda_i}{\lambda_j}\right)-\ln\left(\frac{\varepsilon(\lambda_i)}{\varepsilon(\lambda_j)}\right)}}
\end{equation}

Compared with other radiative thermometry methods that typically require constructing an emissivity model to account for emissivity effects, colorimetric thermometry offers a key advantage when the selected wavelength pair is sufficiently close. Under this condition, the emissivities at the two wavelengths can be reasonably approximated as equal, i.e., $\varepsilon(\lambda_i)\approx\varepsilon(\lambda_j)$, and thus $\varepsilon(\lambda_i)/\varepsilon(\lambda_j)\approx 1$. Accordingly, the emissivity-ratio term in Equation (\ref{tempij_full}) can be omitted, leading to the simplified inversion in Equation (\ref{tempij_final})\cite{qi2024simulation}. This property reduces the sensitivity of the temperature estimation to emissivity uncertainty and constitutes an important advantage of colorimetric thermometry over other radiance-based approaches. 

\begin{equation}
	\label{tempij_final}
	{{\hat T}_{ij}} = \frac{{{c_2} \cdot \left( {\frac{1}{{{\lambda _j}}} - \frac{1}{{{\lambda _i}}}} \right)}}{{\ln \left( {\hat r_{ij}^e} \right) + 5\ln \left( {\frac{{{\lambda _i}}}{{{\lambda _j}}}} \right)}}
\end{equation}

\subsubsection{Temperature calculation}
Once the radiance estimates are obtained, the physical model defined in Equation (\ref{tempij}) is employed to calculate the temperature estimate for each expert.
\begin{equation}
	\label{tempij}
	{{\hat T}_{ij}} = \frac{{{c_2} \cdot \left( {\frac{1}{{{\lambda _j}}} - \frac{1}{{{\lambda _i}}}} \right)}}{{\ln \left( {\hat r_{ij}^r} \right) + 5\ln \left( {\frac{{{\lambda _i}}}{{{\lambda _j}}}} \right)}}
\end{equation}

\subsection{Weighted decision-making}
\subsubsection{Gating network}
\begin{figure*}[htpb]
\centering
    \includegraphics[width=1.00\linewidth]{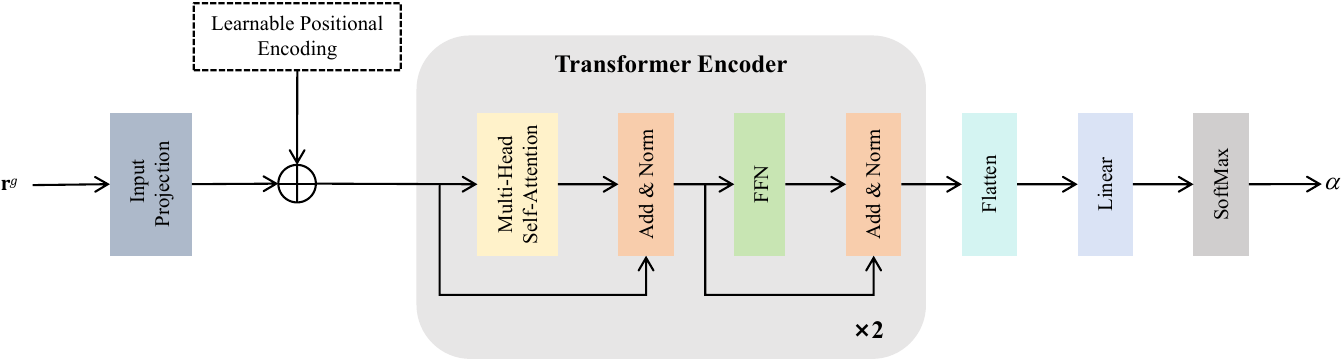}
\caption{The inner architecture of the proposed Transformer-based gating network.}
\label{transformer}
\end{figure*}

The internal architecture of the proposed gating network is illustrated in Fig. \ref{transformer}. It takes six grayscale ratios as input and outputs a weight vector for the six experts. The six grayscale ratios are not a temporal sequence but spectral features computed from different band pairs. Accordingly, the gating network is designed to learn a data-driven expert-weighting function from these ratio features for expert fusion. In our best-performing setting, we adopt a Transformer-based gating network, where each ratio is treated as a token and multi-head self-attention explicitly captures inter-feature interactions. Specifically, the six ratio features are first mapped by an input projection layer to obtain a 64-d token embedding for each ratio. A learnable positional embedding is then added to preserve the identity of each band-pair ratio. The embedded tokens are passed through a 2-layer Transformer encoder, where multi-head self-attention models the interactions among the ratio features. Finally, the encoder outputs are flattened and fed into a linear layer followed by a Softmax function to produce the normalized expert-weight coefficients for dynamic expert fusion.

\subsubsection{Dynamical fusion}
The gating network follows the standard MoE weighting formulation and is used to assign sample-dependent weights to the physically defined wavelength-pair experts according to the current grayscale-ratio observation, as formulated in Equation (\ref{fusedtemp}).

\begin{equation}
\label{fusedtemp}
{{\hat T}_f} = \sum\limits_{i = 1}^3 {\sum\limits_{j = i + 1}^4 {{\alpha _{ij}} \cdot } } {{\hat T}_{ij}}
\end{equation}
where ${\alpha _{ij}}$ denotes the weight coefficient of ${{\hat T}_{ij}}$, which can be obtained by Equation (\ref{alpha}).

\begin{equation}
\label{alpha}
\alpha  = {\Psi _{{{\dot \Theta }_\Psi }}}\left( {{{\bf{r}}^g}} \right)
\end{equation}
where ${\bf{r}}^g$ denotes the vectorized grayscale-ratio set composed of all paired spectral bands. $\Psi(\cdot)$ represents the gating network used for weight estimation, with $\dot{\Theta}$ denoting its optimal parameters.

\subsubsection{Loss function}
$\dot{\Theta}$ in Equation (\ref{alpha}) is optimized by the loss function defined in Equation (\ref{thetapsi}).

\begin{equation}
\label{thetapsi}
{{\dot \Theta }_\Psi } = \arg \mathop {\min }\limits_{{\Theta _\Psi }} \sum\limits_{n = 1}^N {{{\left( {{\Psi _{{\Theta _\Psi }}}\left( {{{\bf{r}}^g}\left( n \right)} \right) \odot {\bf{\hat T}}\left( n \right) - T\left( n \right)} \right)}^2}}
\end{equation}
where the vectorized set of temperature estimiations ${\bf{\hat T}}$ is defined in Equation (\ref{bfT}).

\begin{equation}
\label{bfT}
{\bf{\hat T}}\left( n \right) = \mathop  \cup \limits_{i = 1}^3 \mathop  \cup \limits_{j = i + 1}^4 {{\hat T}_{ij}}\left( n \right)
\end{equation}

\section{Experiments}\label{experiments}
\subsection{Datasets}
\subsubsection{Dataset construction}
\begin{figure}[htpb]
\centering
    \includegraphics[width=1.00\linewidth]{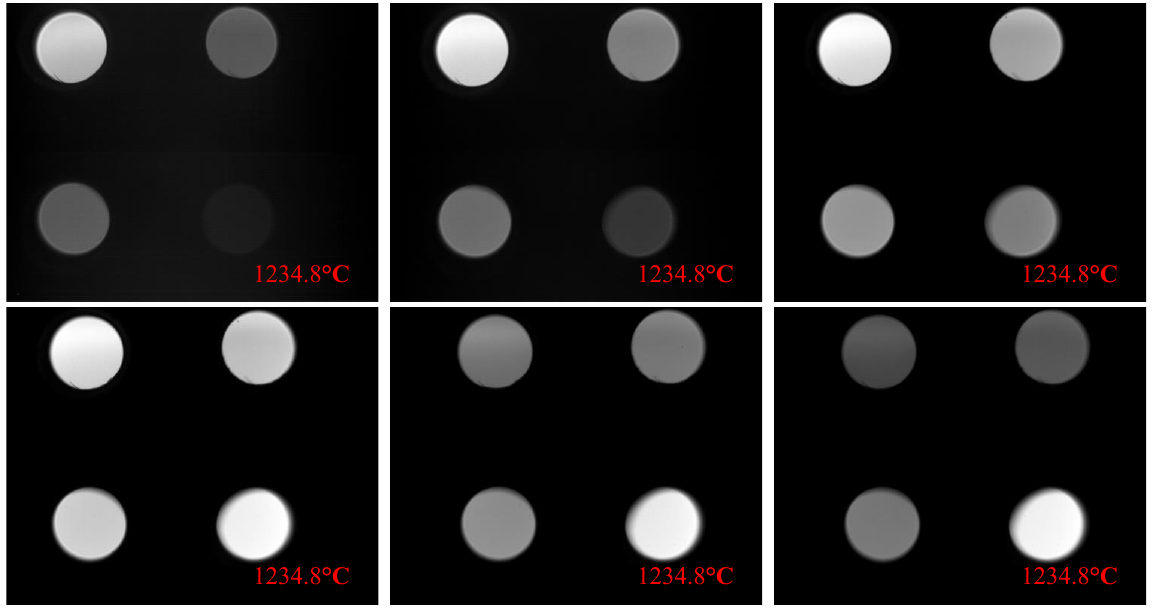}
\caption{Some examples of the constructed dataset.}
\label{dataset}
\end{figure}

The dataset used in this study is collected using an ultra-high-temperature blackbody radiation source and a custom-designed sub-aperture imaging system. Examples of the constructed dataset are shown in Fig. \ref{dataset}.

{\bf{Blackbody:}} The blackbody source used in this study is manufactured by Guangzhou Dongbu Technology Co., Ltd., with the model designation EP3000HY25A1. It features an operational temperature range of 600$\sim$3000 °C, a radiation surface diameter of 25 mm, and an emissivity exceeding 0.999. In addition, the device exhibits excellent temperature uniformity and stability, which are critical for ensuring the reliability of the experimentally collected data.

\begin{figure*}[t]
\centering
    \includegraphics[width=1.00\linewidth]{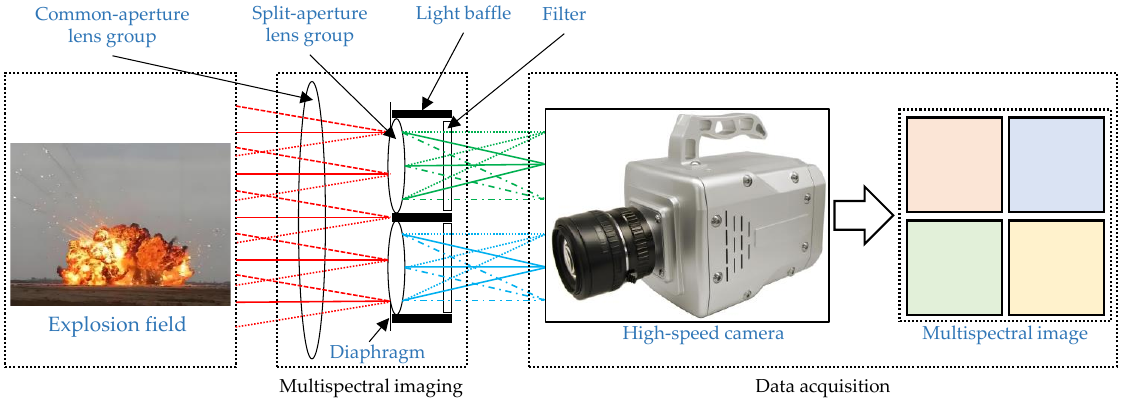}
\caption{Schematic diagram of the split-aperture multispectral imaging system employed in this study, showing its key structural components.}
\label{imagingsystem}
\end{figure*}

{\bf{Imaging system:}}A custom-designed sub-aperture imaging system integrated with a high-speed camera is utilized to capture multi-band images of blackbody radiation. Its hardware architecture is shown in Fig. \ref{imagingsystem}. This imaging system has a focal length of 80 mm, a field of view (FOV) of approximately 4°, and an F-number of 3.8. It is configured with four filter channels, whose central wavelengths are 480 nm, 532 nm, 565 nm, and 632 nm respectively. To avoid image overexposure, each channel is equipped with an appropriate attenuation filter. During calibration, the front end of the lens was positioned 1 m from the blackbody furnace, and focusing was achieved by adjusting the back focal distance. The calibration temperature range spanned from 1194.8 °C to 2955.2 °C.

\begin{figure*}[htpb]
\centering
    \includegraphics[width=1.00\linewidth]{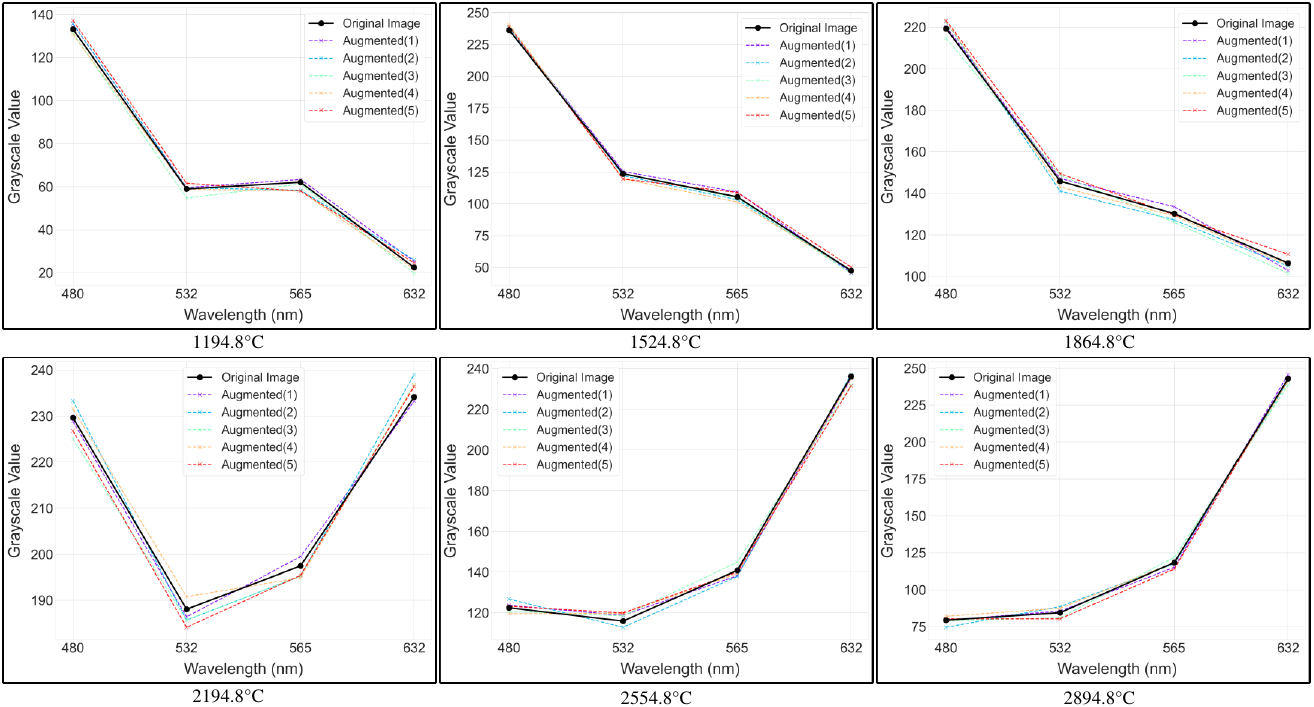}
\caption{Some examples of augmented spectral curves under different channel-wise intensity perturbations.}
\label{aug}
\end{figure*}

{\bf{Acquisition procedure:}} The data acquisition was performed as follows: the blackbody furnace was set to 1194.8 °C. After the temperature stabilized, the upper control computer monitored the image saturation in real time. If saturation occurred, the camera’s exposure time was adjusted to ensure image quality, and then blackbody radiation images were recorded at this temperature. Subsequently, the furnace temperature was increased in 10 °C increments, and the above steps were repeated until reaching 2955.2 °C. To improve measurement accuracy, five images were captured at each temperature point, and their mean grayscale value was used as the final grayscale characteristic for that temperature. In addition, the ground-truth temperature and exposure time are encoded in the image filename.

\begin{table*}[t]
	\centering
	\caption{Model configurations used in our experiments.}
	\label{model_params}
	\setlength{\tabcolsep}{6pt}
	\renewcommand{\arraystretch}{1.15}
	\begin{tabular}{l p{12.2cm}}
		\toprule
		Method & Parameters \\
		\midrule
		MLP-Gated MoE & experts=6; Gate MLP: $6\!\rightarrow\!64\!\rightarrow\!64\!\rightarrow\!6$; ReLU; Softmax over experts \\
		CNN-Gated MoE & experts=6; Gate Conv1d: $1\!\rightarrow\!32$ (k=3, p=1) + ReLU, $32\!\rightarrow\!64$ (k=3, p=1) + ReLU; Output: Linear($64\!\times\!6\!\rightarrow\!6$); Softmax over experts \\
		LSTM-Gated MoE & experts=6; LSTM: input\_dim=1, hidden=64, layers=2, dropout=0.1; Output: Linear($64\!\rightarrow\!6$); Softmax over experts \\
		GRU-Gated MoE & experts=6; GRU: input\_dim=1, hidden=64, layers=2, dropout=0.1; Output: Linear($64\!\rightarrow\!6$); Softmax over experts \\
		Transformer-Gated MoE & experts=6; input\_proj: Linear($1\!\rightarrow\!64$); learnable pos emb: $(1,6,64)$; Encoder: layers=2, $d_{\text{model}}$=64, heads=4, FFN=128, dropout=0.1; Output: Linear($6\!\times\!64\!\rightarrow\!6$); Softmax over experts \\
		\bottomrule
	\end{tabular}
\end{table*}

\subsubsection{Data Augmentation}
To enhance the model’s generalization and stability under different noise conditions, data augmentation is introduced during training. Specifically, we simulate measurement perturbations by directly adding small random offsets to the grayscale intensities of the four extracted wavelength channels. For each sample, the intensity of each channel is independently perturbed by a random value uniformly sampled from $[-5, 5]$, and the resulting values are clipped to the valid range. Based on the perturbed four-channel measurements, each original sample is augmented into five new samples with different random realizations, as shown in Fig. \ref{aug}. This approach significantly expands the training set and encourages the model to learn noise-insensitive features, thereby improving robustness in practical measurements.

\begin{table*}[t] \normalsize \addtolength{\tabcolsep}{-3.5pt}\setlength{\tabcolsep}{5.5pt}
	\renewcommand\arraystretch{1.00}
	\setlength{\abovecaptionskip}{1pt}
	\setlength{\belowcaptionskip}{1pt}
	\newcommand{\tabincell}[2]{\begin{tabular}{@{}#1@{}}#2\end{tabular}}
	\centering
	\caption{The contrasting experimental results of different methods on the constructed explosion filed dataset.}
	\label{tabcontrasting}
	\small\begin{tabular}{|c|c|c|c|}
		\hline
        {Type}                          &{Method}                                                       &{Mean RMSE (K)}                        &{Mean MAE (K)}    \\
        \hline
       {Physical model}  &{Spectral colorimetry \cite{gao2021error}}           &{34.70$\pm$0.00}                           &{27.25$\pm$0.00}\\
        \hline
        \multirow{3}{*}{Classical regressor}
                                           &{SVR \cite{hearst1998support}}                    &{50.44$\pm$0.00}                            &{31.07$\pm$0.00} \\
        \cline{2-4}
		                             &{LWR \cite{cleveland1988locally}}                 &{37.87$\pm$0.00}                             &{29.89$\pm$0.00}  \\
		\cline{2-4}
		&{Global polynomial regression (4th-order)}                 &{26.45$\pm$0.00}                             &{20.04$\pm$0.00}  \\
        \hline
        \multirow{4}{*}{Deep regressor}
                                         &{MLP \cite{rosenblatt1958perceptron}}         &{22.37$\pm$1.77}                             &{17.43$\pm$1.13}   \\
        \cline{2-4}
		                           &{CNN \cite{lecun1989handwritten}}                        &{19.74$\pm$3.31}                    &{15.72$\pm$2.99}   \\
        \cline{2-4}
                                   &{Lightweight Attention \cite{vaswani2017attention}}                        &{16.14$\pm$1.08}                    &{12.17$\pm$0.88}   \\
        \cline{2-4}
		                           &{GraDL \cite{xing2023graphical}}                   &{376.35$\pm$22.43}                          &{285.63$\pm$20.05}   \\
        \hline
        \multirow{5}{*}{Gated MoE}
                                       
		                            &{MLP-Gated MoE}                                           &{20.53$\pm$0.05}                         &{14.73$\pm$0.09}\\
	\cline{2-4}
                                        &{CNN-Gated MoE}                                           &{20.38$\pm$0.06}                          &{14.58$\pm$0.08}\\
      \cline{2-4}
                                        &{LSTM-Gated MoE}                                  &{18.41$\pm$2.62}                          &{12.52$\pm$2.30}   \\
      \cline{2-4}
                                        &{GRU-Gated MoE}                                           &{18.02$\pm$2.58}                          &{12.25$\pm$2.19}   \\
      \cline{2-4}
                                      &{Transformer-Gated MoE}                                                                     &{12.70$\pm$1.32}                                    &{7.96$\pm$0.86} \\
		\hline
        \multicolumn{4}{l}{\footnotesize *Standard deviation is not applicable (N/A) for deterministic methods without training.} \\
       \hline
	\end{tabular}
\end{table*}

\subsection{Evaluation metrics}
To comprehensively and objectively evaluate the performance of the proposed method, we use Mean Absolute Error (MAE) and Root Mean Square Error (RMSE).

The MAE measures the average magnitude of the errors between predicted and true values, defined as the arithmetic mean of the absolute deviations. Since it treats all errors equally regardless of their sign, MAE provides an intuitive assessment of the overall predictive accuracy. Mathematically, MAE is defined as Equation (\ref{mae}).
\begin{equation}
\label{mae}
	\mathrm{MAE} = \frac{1}{N} \sum_{i=1}^{N} \left| \hat{T}_i - T_i \right|,
\end{equation}
where $N$ denotes the number of samples, $\hat{T}_i$ is the predicted temperature for the $i$-th sample, and $T_i$ is the corresponding true temperature.

The RMSE quantifies prediction errors by computing the square root of the mean squared deviations, which is defined as Equation (\ref{rmse}).
\begin{equation}
\label{rmse}
	\mathrm{RMSE} = \sqrt{\frac{1}{N} \sum_{i=1}^{N} \left( \hat{T}_i - T_i \right)^2 }.
\end{equation}

Throughout this study, all temperature values are expressed in Kelvin (K). For each model configuration, 10 independent experiments were conducted. The reported metrics include the mean MAE, mean RMSE, and their standard deviations across these runs, thereby reflecting both the accuracy and performance fluctuations under the current controlled dataset.

\subsection{Experimental settings}
\begin{table*}[ht]
	\centering
	\caption{Validation performance of each expert network with different polynomial degrees (MAE and RMSE in K). Optimal polynomial degree for each expert is highlighted in bold.}
\renewcommand\arraystretch{1.00}
	\label{tabpoly}
	\begin{tabular}{|c|c|cc|cc|cc|cc|cc|}
		\hline
		\multirow{2}{*}{\textbf{Expert No.}}& \multirow{2}{*}{\textbf{$\lambda$ Pair(nm)}} & \multicolumn{2}{c|}{\textbf{2nd}} & \multicolumn{2}{c|}{\textbf{3rd}} & \multicolumn{2}{c|}{\textbf{4th}} & \multicolumn{2}{c|}{\textbf{5th}} & \multicolumn{2}{c|}{\textbf{6th}} \\
		\cline{3-4}                         \cline{5-6}                        \cline{7-8}               \cline{9-10}               \cline{11-12}
		   &{}& MAE     & RMSE       & MAE                    & RMSE                     & MAE                    & RMSE                         & MAE                   & RMSE                          & MAE & RMSE \\
		\hline
		1 &(480,532) & 37.41    & 43.48      & \textbf{30.03}                    & \textbf{37.73}                     & 30.98      & 38.26              & 31.06                 & 37.50                         & 30.81 & 38.04 \\
\hline
		2&(480,565) & 50.93    & 69.81      & \textbf{34.75}                    & \textbf{61.79}                      & 38.93      & 60.44              & 38.46                 & 58.17                         & 38.24 & 58.53 \\
\hline
		3&(480,632) & 148.45  & 176.69    & 126.59                  & 153.14                    & 60.53                   & 72.40                           & 62.27                 & 86.28                          & \textbf{45.66} & \textbf{66.61} \\
\hline
		4&(532,565) & 189.17  & 236.25    & 178.98     & 224.52      & 178.62                 & 225.11                         & \textbf{172.07}               & \textbf{224.51}                      & 191.30 & 233.07 \\
\hline
		5&(532,632) & 81.42    & 93.84      & 49.57                    & 61.64                      & \textbf{25.48}                   & \textbf{38.20}                          & 26.66    & 39.51             & 27.05 & 39.89 \\
\hline
		6&(565,632) & 86.16    & 105.68      & 48.38                    & 59.93                      & 23.45                   & 36.90                           & 21.78                 & 26.57                          & \textbf{20.35}             & \textbf{25.19} \\
		\hline
	\end{tabular}
\end{table*}

\begin{figure*}[htpb]
\centering
    \includegraphics[width=1.00\linewidth]{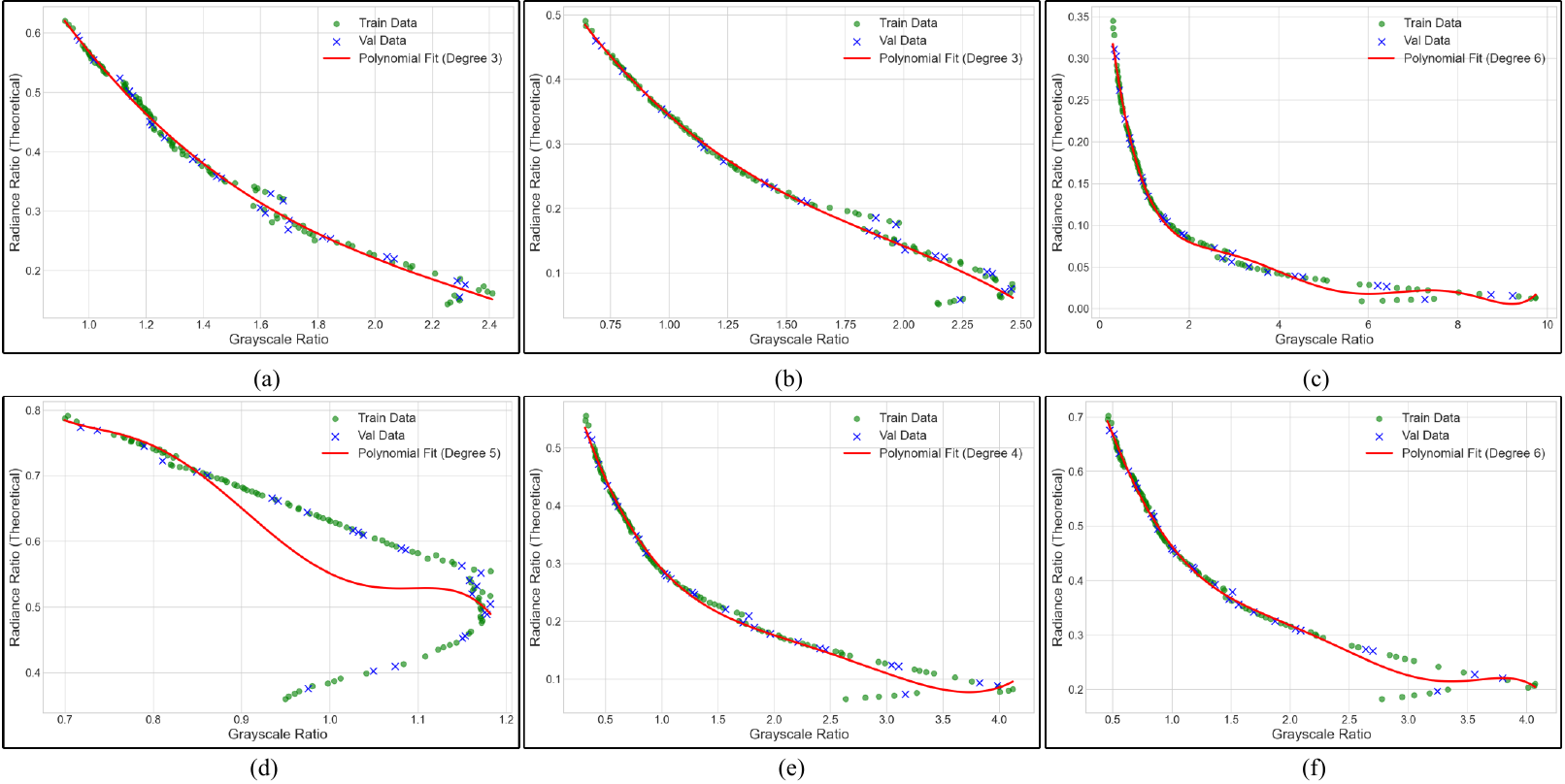}
\caption{The visualizations of the optimal polynomial functions correspond to different wavelength pairs, where (a)-(f) denote the curves of first to sixth experts respectively. The polynomial degree (2--6) for each expert is selected by minimizing the validation MAE/RMSE, and the corresponding quantitative errors and optimal degrees are reported in Table\ref{tabpoly}.}.
\label{poly}
\end{figure*}

\begin{figure}[htpb]
\centering
    \includegraphics[width=1.00\linewidth]{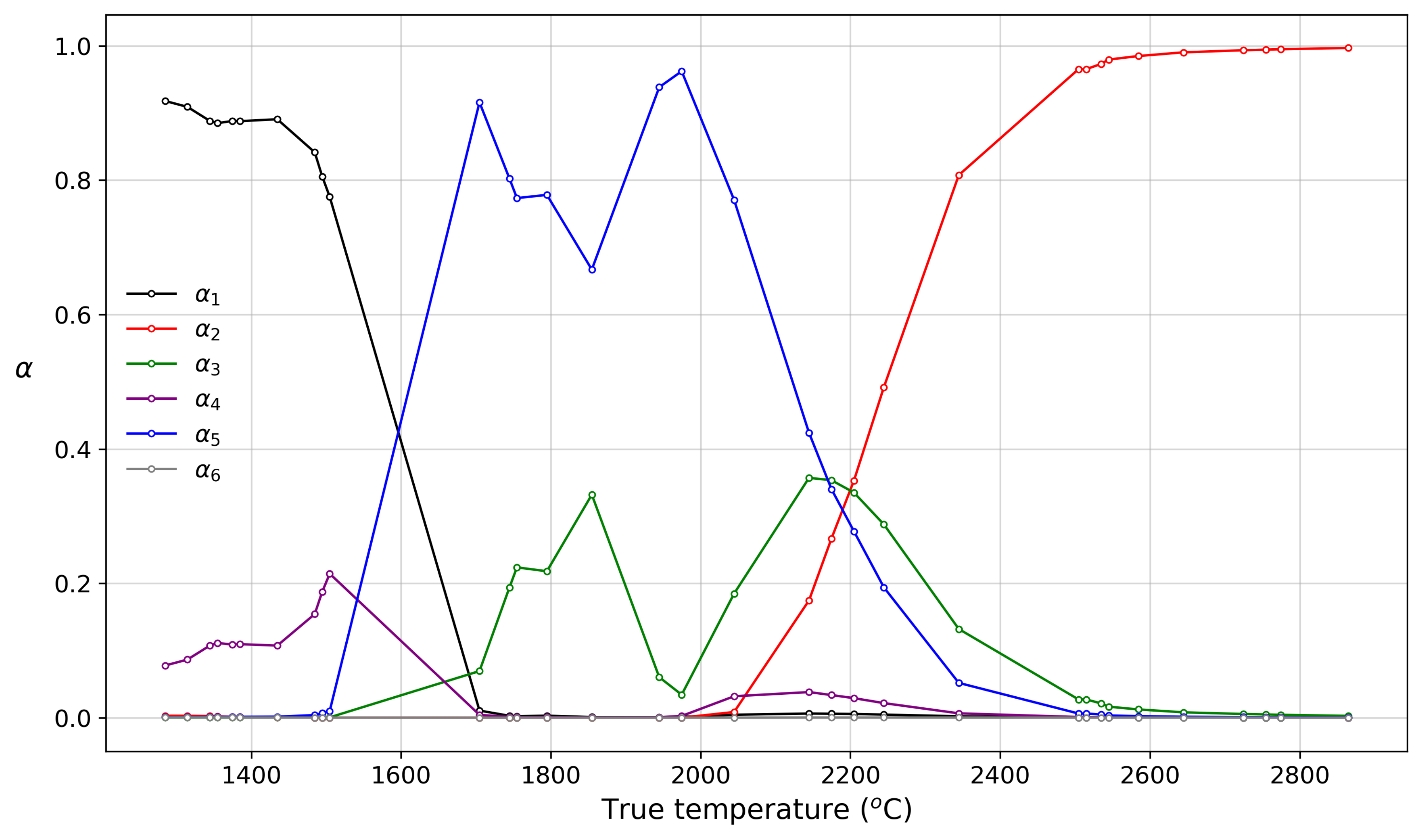}
\caption{Weight coefficients of different experts \emph{vs.} temperature.}
\label{weights}
\end{figure}

\begin{figure}[htpb]
	\centering
	\includegraphics[width=1.00\linewidth]{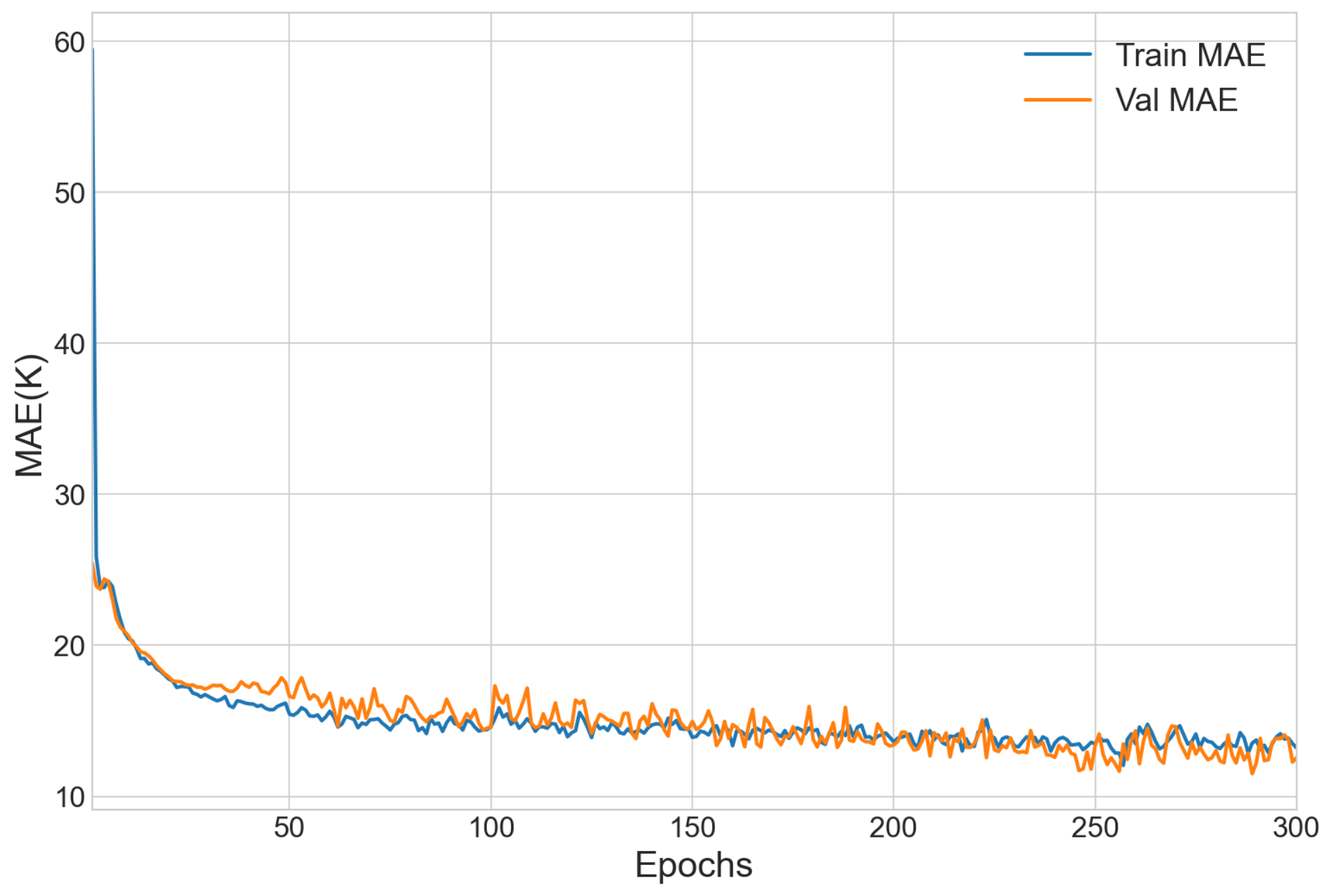}
	\caption{Training and validation MAE curves of the proposed model.}
	\label{train_curve}
\end{figure}

To ensure the validity of the experimental evaluation, this section details the experimental setup, including the implementation details, YOLOv8 settings, data partitioning, model architecture and expert configuration, gating network design, training strategy, and the evaluation protocol.

\subsubsection{Implementation Details}
Experiments were run on a Windows 10 workstation equipped with an NVIDIA GeForce RTX 3090 GPU and Intel Core i9-12900K CPU. Model implementation and training used the PyTorch deep learning framework (version 1.12.1).

\subsubsection{YOLOv8 setting}
 We fine-tuned Ultralytics YOLOv8s (initialized from the pretrained yolov8s.pt) on a single-class dataset for blackbody localization in each spectral-band image. The detector was trained for 50 epochs with an input size of $640\times640$ and a batch size of 4.

\subsubsection{Data partitioning} 
The dataset was randomly split (60\% training, 20\% validation, 20\% testing). For objective evaluation, the split was performed at the temperature-point level by holding out several discrete temperature points for the validation and test sets, ensuring these temperature points were not encountered during training.

\subsubsection{Model Architecture and Expert Configuration}

Colorimetric thermometry estimates temperature from the radiance ratio between two wavelengths. Since our spectral images contain four bands (480, 532, 565, 632\,nm), we form all pairwise combinations ($C_{4}^{2}=6$) and thus build six experts, each performing temperature inversion using one wavelength pair: (480, 532) nm, (480, 565) nm, (480, 632) nm, (532, 565) nm, (532, 632) nm, and (565, 632) nm.
\subsubsection{Gating Network Architecture and Configuration}
To ensure reproducibility and avoid ambiguity in architectural choices, we summarize the detailed configurations of the gating networks for the compared MoE variants in Table \ref{model_params}. Specifically, all MoE variants employ the same six physics-based experts corresponding to the six spectral-band ratio inputs, while only the gating network differs across variants (MLP-, CNN-, LSTM-, GRU-, and Transformer-based gating). For each gating design, Table \ref{model_params} reports the layer-wise structure and key hyperparameters (e.g., hidden dimension, number of layers/heads, and dropout).

\subsubsection{Training Strategy}
The Adam optimizer is used for parameter optimization, with a learning rate of 0.0005. The model is trained for 300 epochs with a batch size of 16. At the start of each epoch, the training set is shuffled to enhance generalization. The training objective is the L1 loss (mean absolute error) between the predicted and ground-truth temperatures.

\subsubsection{Evaluation Protocol}
All methods were evaluated under the same data partitioning and training settings to ensure a fair comparison.

\subsection{Contrasting experiments}
To comprehensively evaluate the performance of the proposed Gated Mixture of Experts (MoE) model, we conducted rigorous comparisons against a set of baseline methods. These baselines range from traditional single-channel approaches to MoE models employing other mainstream neural network architectures. The goal of this section is to systematically evaluate the proposed method through multi-dimensional performance metrics.

Table \ref{tabcontrasting} summarizes the results of all methods under identical experimental conditions, where each method was independently trained and evaluated 10 times. We report the mean RMSE, the standard deviation (Std Dev), and the best RMSE among the runs.

\subsubsection{Contrasting methods}
\begin{itemize}
	\item \textbf{Spectral colorimetry} represents the traditional dual-wavelength temperature measurement approach, where only the single wavelength pair with the best calibration performance is used for temperature estimation. It serves as a physics-based reference baseline for comparison with the proposed method.
       \item \textbf{Classical regressors} such as SVR (Support Vector Regression), LWR (Locally Weighted Regression), and a fourth-order global polynomial regression model, which all estimate temperature directly in a data-driven manner.
        \item \textbf{Deep regressors} (e.g., MLP (Multi-Layer Perceptron), CNN (Convolutional Neural Network), GraDL (Graph Deep Learning), and Lightweight Attention) refer to methods that use deep neural networks to directly estimate temperature values from the input grayscale-ratio features, without explicit physical experts or fusion mechanisms.
    
	\item \textbf{Gated MoE} employs a standard Multi-Layer Perceptron (MLP), Convolutional Neural Network (CNN), Gated Recurrent Unit (GRU), Long Short-Term Memory (LSTM), or Transformer as its gating network to adaptively assign weights to the experts.
\end{itemize}
\begin{figure*}[t]
	\centering
	\includegraphics[width=1.00\linewidth]{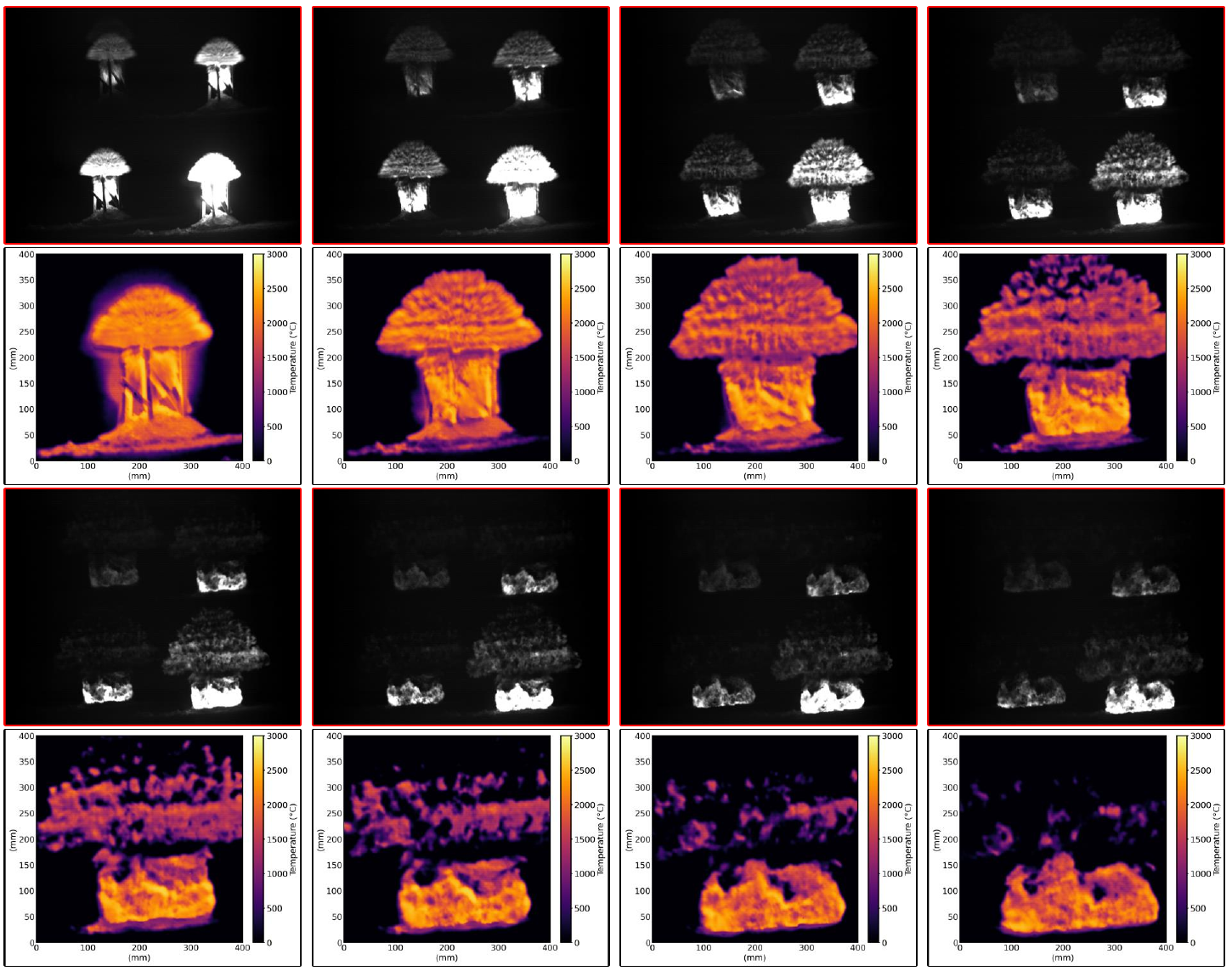}
	\caption{The visualizations of the proposed method on practical two-dimensional explosion field data.}
	\label{pravis}
\end{figure*}

\begin{figure*}[t]
	\centering
	\includegraphics[width=1.00\linewidth]{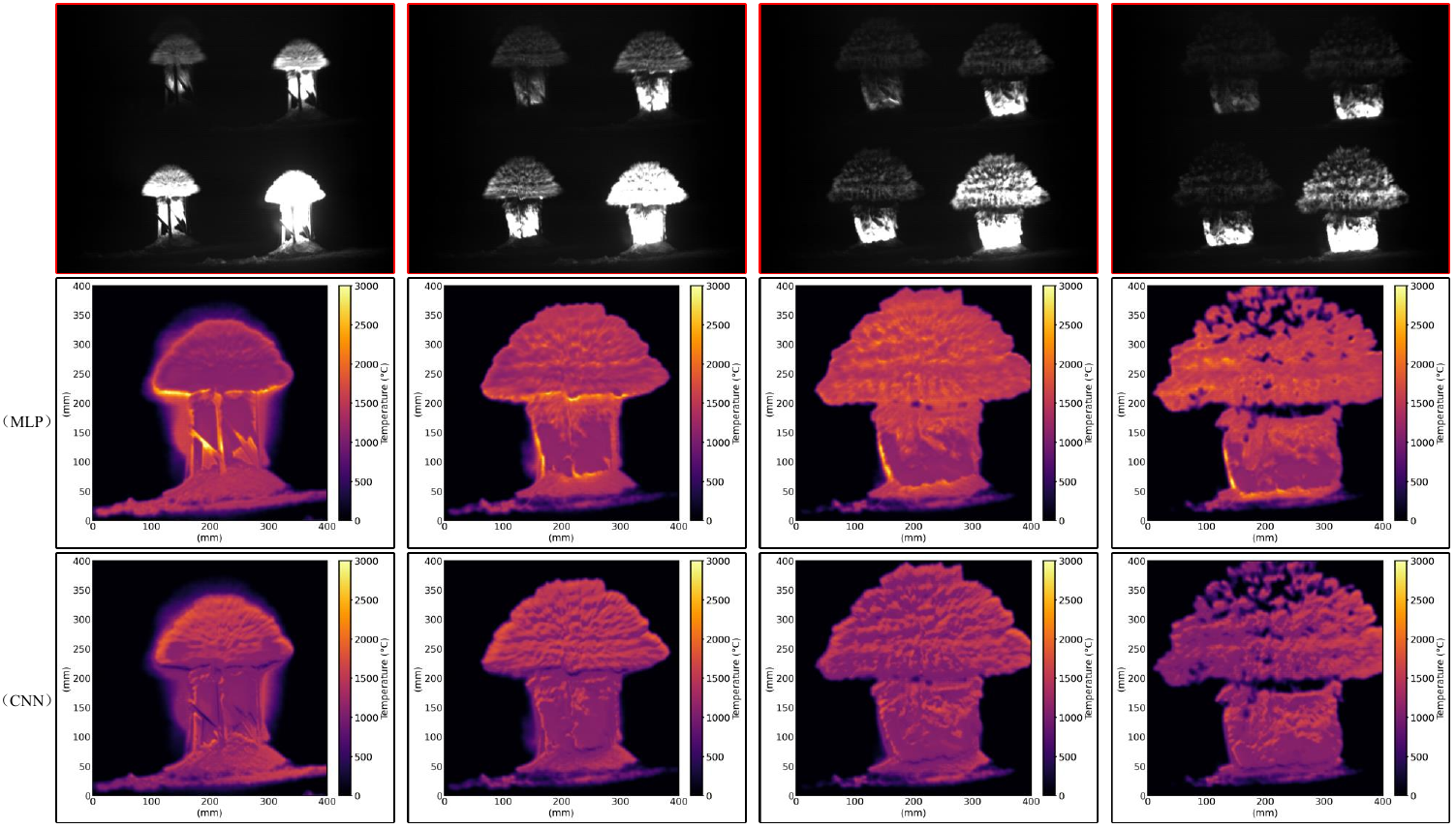}
	\caption{The visualizations of baseline methods on practical two-dimensional explosion field data.}
	\label{pravis_comp}
\end{figure*}

\subsubsection{Analysis of Results}
From the results in Table~\ref{tabcontrasting}, the following conclusions can be drawn. GraDL is included as a recent deep-learning baseline for multispectral radiometric temperature inversion. All baselines are evaluated under the same data partitioning, preprocessing procedure, evaluation metrics, and repeated-run protocol. The relatively large error of GraDL may be attributed to its purely data-driven nature, as it does not explicitly incorporate the calibrated grayscale-ratio-to-radiance-ratio mapping or the physical inversion process, which may make it more sensitive to the limited-size dataset and wavelength-pair-dependent nonlinearities.

\begin{itemize}
	\item \textbf{Superiority of the MoE framework}: All MoE-based models significantly outperform spectral colorimetry in terms of Mean RMSE and MAE, indicating the effectiveness of adaptive multi-expert fusion over the conventional single-expert temperature estimation strategy.
	\item \textbf{Leading average accuracy}: Among the five gating mechanisms, the proposed Transformer-Gated MoE model achieves the best overall average performance in terms of Mean RMSE and MAE, indicating that the self-attention-based gating strategy is effective in capturing inter-spectral relationships for temperature estimation.
	\item \textbf{Stability}: As presented in Table~\ref{tabcontrasting}, the MoE-based methods generally exhibit relatively small standard deviations across repeated experiments, indicating improved stability under the current evaluation setting. This consistency suggests that the advantage of GMoE extends beyond improving global average accuracy to providing more stable estimation across different temperature intervals. Given that individual experts possess distinct error profiles across the temperature spectrum, adaptive fusion helps mitigate local performance degradation, thereby enhancing the overall reliability of the model.
\end{itemize}

\begin{figure*}[htpb]
	\centering
	\includegraphics[width=1.00\linewidth]{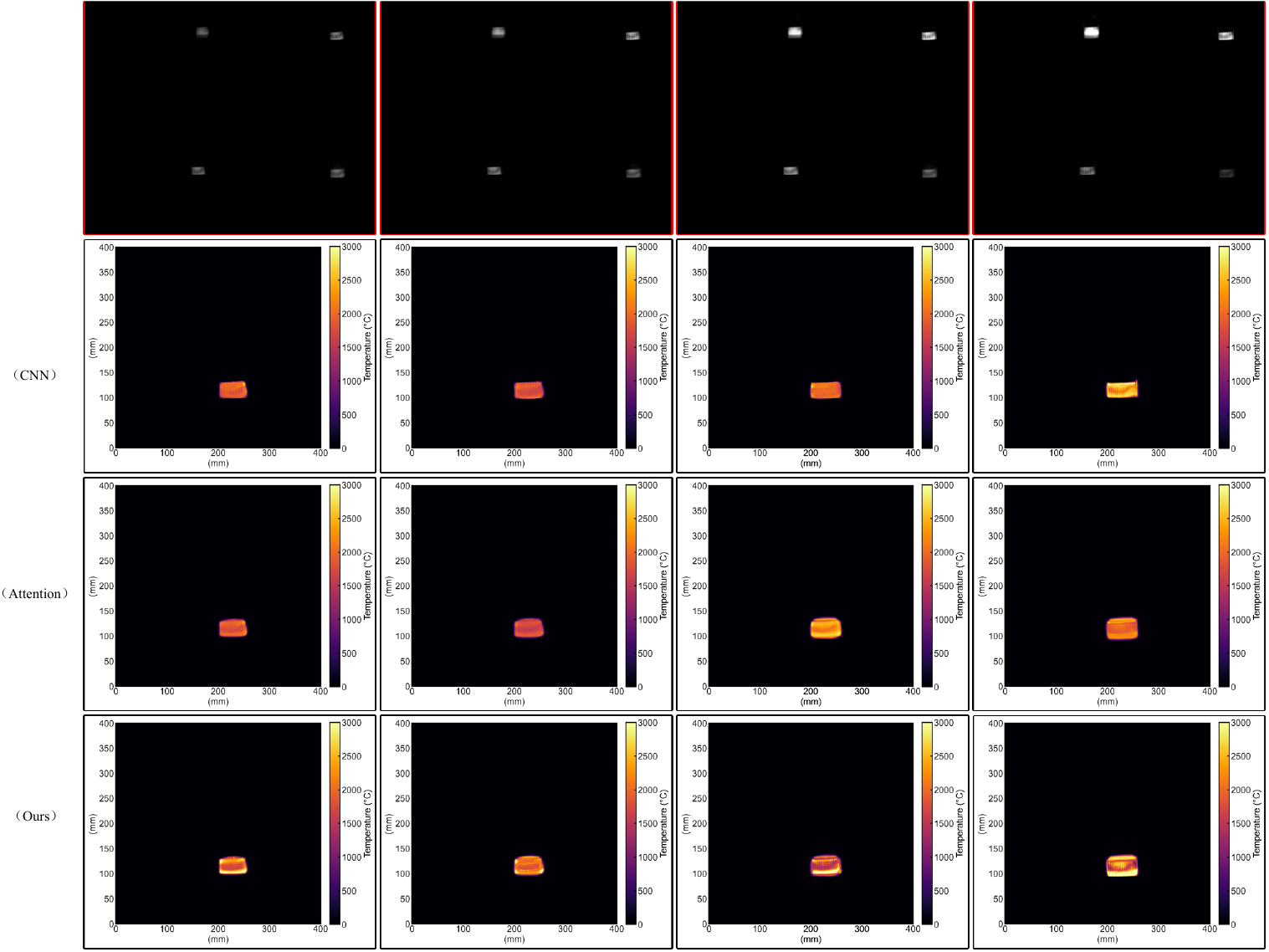}
	\caption{Visualization comparison on representative unseen two-dimensional halogen-lamp data. The first row shows the input images, and the following rows show the reconstructed temperature maps obtained by CNN, Lightweight Attention, and Transformer-Gated MoE, respectively.}
	\label{lusudeng}
\end{figure*}

\begin{table*}[t]
	\centering
	\caption{Noise-robustness stress tests under additive intensity perturbations on spectral bands.}
	\label{noise_robust}
	\setlength{\tabcolsep}{6pt}
	\renewcommand{\arraystretch}{1.15}
	\begin{tabular}{|c|c|c|}
		\hline
		Method & Mean RMSE (K) & Mean MAE (K) \\
		\hline
		Transformer-Gated MoE $\rightarrow$ noise-perturbed test set & 36.90$\pm$2.45 & 26.24$\pm$0.86 \\
		\hline
		Transformer-Gated MoE+Aug $\rightarrow$ noise-perturbed test set & 26.56$\pm$1.71 & 20.26$\pm$0.99 \\
		\hline
	\end{tabular}
\end{table*}

\begin{table*}[t]
	\centering
	\caption{Ablation study on ROI sampling strategies.}
	\label{roi_ablation}
	\setlength{\tabcolsep}{6pt}
	\renewcommand{\arraystretch}{1.15}
	\begin{tabular}{|c|c|c|}
		\hline
		\textbf{Method} & \textbf{Mean RMSE (K)} & \textbf{Mean MAE (K)} \\
		\hline
		Transformer-Gated MoE + Single-point ROI & 16.22$\pm$1.58 & 10.84$\pm$1.47 \\
		\hline
		Transformer-Gated MoE + MPROI            & 12.70$\pm$1.32 & 7.96$\pm$0.86 \\
		\hline
	\end{tabular}
\end{table*}

\begin{table*}[t]
	\centering
	\caption{Extrapolation evaluation (Train: 1000--2350 K, Val: 2350--2600 K, Test: 2600--3000 K).}
	\label{extra_test}
	\setlength{\tabcolsep}{6pt} 
	\renewcommand{\arraystretch}{1.15} 
	\begin{tabular}{|c|c|c|}
		\hline
		Method & Mean RMSE (K) & Mean MAE (K) \\
		\hline
		Spectral colorimetry & 80.72$\pm$0.00 & 78.78$\pm$0.00 \\
		\hline
		Transformer-Gated GMoE  & 55.46$\pm$26.42 & 53.00$\pm$27.00 \\
		\hline
	\end{tabular}
\end{table*}

\begin{table*}[t]
	\centering
	\caption{Comparison of different expert fusion strategies.}
	\label{fusion_strategy}
	\setlength{\tabcolsep}{6pt}
	\renewcommand{\arraystretch}{1.15}
	\begin{tabular}{|c|c|c|}
		\hline
		Method & Mean RMSE (K) & Mean MAE (K) \\
		\hline
		Transformer-Gated MoE (Top-1 hard selection) & 22.91$\pm$5.10 & 17.29$\pm$2.53 \\
		\hline
		fixed-weight fusion (1/6 equal weights) & 108.34$\pm$0.00 & 102.72$\pm$0.00 \\
		\hline
		Transformer-Gated MoE (soft weighted fusion) & 12.70$\pm$1.32 & 7.96$\pm$0.86 \\
		\hline
	\end{tabular}
\end{table*}
\begin{table*}[t]
	\centering
	\caption{Quantitative comparison on the unseen halogen-lamp dataset.}
	\label{halogen_quantitative}
	\setlength{\tabcolsep}{6pt}
	\renewcommand{\arraystretch}{1.15}
	\begin{tabular}{|c|c|c|}
		\hline
		Method & MAE (K) & RMSE (K) \\
		\hline
		CNN & 412.30 & 461.43 \\
		\hline
		Lightweight Attention & 417.90 & 620.25 \\
		\hline
		Transformer-Gated MoE & \textbf{268.81} & \textbf{302.21} \\
		\hline
	\end{tabular}
\end{table*}
\subsection{Ablation Study}
\subsubsection{Sensitivity to Intensity Perturbations}
To evaluate the effect of the proposed data augmentation strategy, we conduct a preliminary sensitivity analysis under simulated channel-intensity perturbations, as summarized in Table~\ref{noise_robust}. Two variants, Transformer-Gated MoE and Transformer-Gated MoE+Aug, are evaluated on the same perturbed test set with all other configurations kept identical. The augmented model achieves better performance, suggesting improved tolerance to moderate input intensity fluctuations. This experiment should be regarded as a preliminary evaluation rather than a comprehensive robustness validation under realistic explosion-field disturbances.
\subsubsection{Effect of ROI Sampling Strategy}
To study the effect of ROI sampling, we compare a simplified single-point ROI baseline, in which the channel measurement is obtained by directly averaging all pixel intensities within the ROI region $R$, with the previously introduced MPROI strategy, while keeping all other settings unchanged. As reported in Table \ref{roi_ablation}, MPROI yields consistently lower estimation errors than single-point ROI averaging, suggesting that aggregating multiple ROI measurements provides more stable band observations under intra-ROI intensity variations.

\subsubsection{Extrapolation Evaluation Across Temperature Ranges}
Polynomial fitting can be sensitive when applied beyond the calibration range. To examine the extrapolation performance toward higher temperatures, we conduct a range-split evaluation: samples within $1000$--$2350~^{\circ}\mathrm{C}$ are used for training, $2350$--$2600~^{\circ}\mathrm{C}$ for validation, and $2600$--$3000~^{\circ}\mathrm{C}$ for testing. As summarized in Table~\ref{extra_test}, the proposed Transformer-Gated GMoE achieves a mean MAE of $53.00$ K and a mean RMSE of $55.46$ K on the extrapolation test set. Compared with spectral colorimetry, which obtains an MAE of $78.78$ K and an RMSE of $80.72$ K, the proposed framework yields lower errors under this out-of-range setting. Nevertheless, the extrapolation errors are evidently higher than those observed within the calibrated range, accompanied by relatively large standard deviations. This indicates that the out-of-range stability of the proposed framework remains limited. Therefore, further validation with broader calibration data is still needed to assess its generalization capability beyond the current calibration range.

\subsubsection{Effect of Expert Fusion Strategy}
To further evaluate the role of the expert fusion strategy, three variants are compared, namely, Top-1 hard selection, fixed-weight fusion, and the proposed soft weighted fusion. In the Top-1 hard-selection variant, only the expert associated with the largest gating weight is retained for the final prediction. In the fixed-weight fusion variant, the outputs of the six experts are combined using predefined input-independent equal weights, i.e., $w_i=1/6$ for $i=1,\ldots,6$. All other settings are kept unchanged.
As reported in Table \ref{fusion_strategy}, the proposed soft weighted fusion achieves the lowest Mean RMSE and Mean MAE among the three variants. Compared with Top-1 hard selection, it avoids relying exclusively on a single expert. Compared with fixed-weight fusion, it further benefits from input-dependent adaptive weighting of expert contributions. These results indicate that the performance improvement arises not only from multi-expert fusion itself, but also from the adaptive expert-allocation capability of the gating mechanism.

\subsubsection{Effect of Model Capacity}
To examine whether the improvement is simply due to increased model capacity, we further introduce a capacity-matched single-expert Transformer regressor as an additional baseline. This model keeps a comparable Transformer backbone and trainable complexity, but removes the MoE-based adaptive expert fusion. Under the same experimental setting, the single-expert baseline obtains a Mean RMSE of $25.42 \pm 10.53$ K and a Mean MAE of $17.53 \pm 6.24$ K, both of which are higher than those of the proposed GMoE reported above. This result indicates that the performance gain is not merely brought by model capacity, but mainly benefits from the adaptive mixture-of-experts formulation.

\subsection{Detail analysis}
\subsubsection{Analysis of the configuration for expert}
With the wavelength-pair experts fixed, we determine the polynomial degree of each expert via hyperparameter experiments, using validation performance as the selection criterion. As summarized in Table~\ref{tabpoly} and Fig. \ref{poly}, the optimal degree varies across wavelength pairs. Specifically, for each pair we conduct a grid search over candidate degrees (2nd--6th): the polynomial is fitted on the training split and evaluated on the validation split, and the degree achieving the minimum validation error is selected. This selection strategy is physically motivated, since the polynomial in each expert serves as a calibration mapping from the measured grayscale ratio to the theoretical radiance ratio rather than a black-box regressor. Prior multispectral colorimetric thermometry studies report that the grayscale-ratio--to--radiance-ratio relationship exhibits a clear nonlinear trend, and polynomial fitting is commonly adopted to absorb system-dependent factors such as ambient-light interference, filter transmittance, and detector response~\cite{zhao2025}. The use of polynomial regression follows these prior studies and provides a smooth empirical calibration mapping with relatively few fitting parameters, which is suitable for the limited calibration dataset used in this work. Consequently, wavelength-pair-dependent curve characteristics lead to different required model complexities: pairs with more pronounced curvature benefit from higher-degree polynomials, whereas others are well approximated by lower-degree models. Nevertheless, for wavelength pairs with weaker sensitivity or lower calibration reliability in certain temperature ranges, higher-order polynomial fitting may introduce local oscillation and boundary instability. Therefore, each fitted expert is first selected according to validation errors, and the gating network further adaptively adjusts its contribution during fusion, which helps reduce the influence of less reliable wavelength-pair estimates on the final temperature prediction.
\subsubsection{Analysis the weight coefficients for different experts}
{The learned weight coefficients of different experts over the temperature range are presented in Fig.~\ref{weights}. These curves provide further physical insight into the expert-weight allocation mechanism of the proposed gating network. In ratio-based pyrometry, the temperature-estimation reliability of a wavelength pair is not constant over the entire temperature range, but is affected by several wavelength-dependent factors, including radiance-ratio sensitivity, detector response, measurement uncertainty, and emissivity-ratio variation. Therefore, different wavelength pairs may exhibit different levels of reliability in different temperature intervals. This observation is consistent with the wavelength-selection principle reported by Huang et al.~\cite{huang2018new}, which indicates that the two-color radiance-ratio mapping and the corresponding temperature sensitivity are strongly dependent on the selected wavelength pair. As shown in Fig.~\ref{weights}, the gating network does not assign fixed weights to the six experts. Instead, it adaptively redistributes the expert weights as the temperature changes. Specifically, the Mixture of Experts (MoE) model assigns a larger weight to the first expert, corresponding to the $480/532~\mathrm{nm}$ wavelength pair, when the temperature is below $1500~^{\circ}\mathrm{C}$, whereas the second expert, corresponding to the $480/565~\mathrm{nm}$ wavelength pair, becomes more dominant when the temperature exceeds $2500~^{\circ}\mathrm{C}$. This temperature-dependent weight allocation suggests that the gating network tends to emphasize the wavelength-pair expert with higher effective reliability in the corresponding temperature range. Therefore, the learned gating behavior is not merely a data-driven fusion result, but is also physically consistent with the wavelength-dependent sensitivity characteristics of two-color pyrometry.}
\subsubsection{Convergence and generalization analysis}
Fig. \ref{train_curve} reports the training and validation MAE curves of the proposed model. Both curves drop rapidly in the early stage and then decrease gradually, indicating stable convergence. More importantly, the validation MAE closely follows the training MAE throughout the optimization, and no persistent divergence is observed, which is a typical symptom of overfitting. The small oscillations on the validation curve are expected under limited validation data and measurement noise, yet the overall trend remains consistent with the training curve. These observations suggest that the proposed model maintains a comparable fitting behavior on unseen samples and does not exhibit evident overfitting under the current setting.
\subsubsection{Computational Complexity Analysis}
As the proposed GMoE framework is developed for offline analysis of explosion-field data, real-time deployment is not the primary consideration in this work. It should be noted that the reported parameter count and FLOPs refer only to the GMoE estimation module after the multispectral grayscale-ratio features are obtained, rather than the entire end-to-end pipeline. Specifically, the proposed network contains approximately 36.1K parameters and requires 205.4K FLOPs per inference, indicating that the expert-fusion model itself is lightweight. The complete pipeline also involves spectral-channel detection, image alignment, ROI sampling, and pixel-wise temperature reconstruction, whose runtime depends on image resolution and implementation details. Therefore, the reported complexity should be interpreted as the computational cost of the GMoE module rather than the total runtime of the full temperature-reconstruction process.

\subsubsection{Uncertainty Analysis}
Predictive uncertainty is evaluated on the test set using a deep ensemble of 10 independently trained models. For each sample, the mean prediction is taken as the final estimate, and the standard deviation is used to quantify uncertainty. On the test set, the proposed method yields an average uncertainty of 3.93 K and a maximum uncertainty of 12.88 K, with 95\% of the ground-truth temperatures covered by the estimated interval. The same ensemble strategy is further applied pixel-wise to explosion-field images without ground-truth labels. Within the valid temperature range of 1000--3000 $^\circ$C, the global average uncertainty is 13.96 K, with a corresponding 95\% error bound of $\pm$27.37 K. These results indicate that the proposed framework can provide quantitative uncertainty estimates for explosion-field analysis even in the absence of ground-truth labels.

\subsection{Supplementary experiments}

The visualized results of the proposed method applied to practical explosion-field images are presented in Fig.~\ref{pravis}. Specifically, since the input of our method is defined as a spectral vector, we first decompose the multispectral image into a set of $W \times H$ independent spectral vectors (where $W$ and $H$ denote the width and height of the image, respectively). Subsequently, the temperature value corresponding to each individual spectral vector is estimated using the proposed model. Finally, these estimated temperature values are aggregated and reconstructed to generate a two-dimensional temperature distribution map. Although no ground-truth temperature is available for these practical explosion-field images, the accuracy of the estimated temperature maps cannot be quantified using numerical indicators. Instead, the uniformity of spatial distribution and the consistency of temporal variation are used as qualitative cues to assess whether the reconstructed maps exhibit physically plausible behavior, rather than as a quantitative validation. In future work, controlled experiments with synchronized reference measurements will be conducted to enable quantitative validation.

To enable a more direct qualitative comparison on practical explosion-field images, we additionally provide baseline visualizations on the same inputs in Fig. \ref{pravis_comp}. As can be visually observed by comparing Fig. \ref{pravis_comp} with Fig. \ref{pravis}, the baseline results tend to show more local fluctuations and less consistent spatial patterns in some regions, particularly near plume boundaries and in low-radiance areas. By contrast, the proposed method yields temperature maps that appear relatively smoother and spatially more coherent while maintaining the main high-temperature regions. This qualitative comparison offers additional, albeit non-quantitative, evidence that our method can improve the visual plausibility of the reconstructed temperature fields on explosion-field data.

To further examine the performance of the proposed model beyond the blackbody-furnace training data, we apply it to an unseen halogen-lamp image set with available reference temperatures. The inference pipeline is identical to that used for the explosion-field images, \emph{i.e.}, each pixel is treated as an independent spectral vector and the estimated temperatures are reconstructed into a 2-D map. Four representative samples with reference temperatures of $2042$, $2200$, $2400$, and $2856~^\circ$C are visualized in Fig.~\ref{lusudeng}, where the corresponding average predicted temperatures of the proposed Transformer-Gated MoE are $2181.59$, $2102.89$, $2226.73$, and $2455.89~^\circ$C, respectively. For comparison, the reconstructed temperature maps obtained by CNN and Lightweight Attention are also shown in Fig.~\ref{lusudeng}. In addition, the average predicted temperature of the high-temperature region is calculated for each halogen-lamp image and compared with the corresponding reference temperature. The resulting MAE and RMSE are reported in Table~\ref{halogen_quantitative}, where the proposed Transformer-Gated MoE yields lower MAE and RMSE than the CNN and Lightweight Attention baselines on this halogen-lamp dataset. These results provide a quantitative supplement to the visualization results and further support the evaluation of the proposed method on unseen practical radiation data.

\section{Conclusion}\label{conclusion}
In this work, we propose a gated mixture of experts (GMoE) framework that integrates physics-driven and data-driven strategies. This framework is designed to simultaneously address two critical challenges in temperature estimation: the insufficient fitting capability of physical models and the overfitting risk of deep neural networks. Specifically, it decomposes the complex temperature estimation task into multiple expert-specific subtasks. Each expert is developed based on physically calibrated polynomial functions, which model the specific relationship between grayscale ratios and radiance ratios for a given pair of spectral bands. After establishing this relationship, the temperature is derived using a physical model that incorporates the corresponding parameters of each expert. Furthermore, a Transformer-based gating network is adopted to dynamically weight the output of each expert. This weighted fusion ultimately generates the final temperature estimation result. Quantitative experiments on controlled radiative sources demonstrate that the proposed method achieves improved estimation performance under the evaluated settings, while qualitative observations on explosion-field imagery indicate that it can produce physically consistent temperature estimation results in explosion scenarios. These results suggest that the proposed method provides a promising framework for multispectral temperature measurement under the currently evaluated conditions. However, the current study still has limitations. Owing to the limited size of the controlled dataset, the statistical support remains limited, although 10 independent experiments were conducted for each model configuration. Moreover, the increased extrapolation errors, accompanied by relatively large standard deviations, indicate that the out-of-range stability of the proposed framework remains limited. Explosion-field data do not yet support fully quantitative analysis and lack dense and reliable ground-truth verification, which limits rigorous absolute-error assessment. Accordingly, the explosion-field results presented in this work should be regarded primarily as qualitative validation and evidence of physical consistency, rather than strict quantitative evaluation. Future work will focus on expanding the calibrated and explosion-field datasets, improving out-of-range generalization, conducting rigorous validation in realistic explosion scenarios, further evaluating robustness beyond simulated channel-intensity perturbations against practical disturbances such as emissivity variations, smoke/scattering attenuation, registration errors, saturation, and exposure inconsistency, and realizing 3D temperature-field reconstruction.

\bibliographystyle{IEEEtran}
\bibliography{jas}

@article{wang2023development,
  title={The development of a multispectral pyrometer for achievable true temperature field measurements of the explosion flame},
  author={Wang, Zhentao and Dai, Jingmin and Yang, Sen},
  journal={Meas. Sci. Technol.},
  volume={34},
  number={6},
  pages={065501},
  month={Mar.},
  year={2023}
}

@inproceedings{huang2025ders,
  title={Ders: Towards extremely efficient upcycled mixture-of-experts models},
  author={Huang, Yongqi and Ye, Peng and Huang, Chenyu and Cao, Jianjian and Zhang, Lin and Li, Baopu and Yu, Gang and Chen, Tao},
  booktitle={Proc. IEEE/CVF Conf. Comput. Vis. Pattern Recognit. (CVPR)},
  address={Nashville, TN, USA},
  pages={10056--10066},
  year={2025}
}

@inproceedings{wang2025decoupled,
  title={Decoupled feature-based mixture of experts for multi-modal object re-identification},
  author={Wang, Yuhao and Liu, Yang and Zheng, Aihua and Zhang, Pingping},
  booktitle={Proc. AAAI Conf. Artif. Intell.},
  volume={39},
  number={8},
  address={Philadelphia, PA, USA},
  pages={8141--8149},
  year={2025}
}

@inproceedings{rajbhandari2022deepspeed,
  title={Deepspeed-moe: Advancing mixture-of-experts inference and training to power next-generation ai scale},
  author={Rajbhandari, Samyam and Li, Conglong and Yao, Zhewei and Zhang, Minjia and Aminabadi, Reza Yazdani and Awan, Ammar Ahmad and Rasley, Jeff and He, Yuxiong},
  booktitle={Proc. Int. Conf. Mach. Learn.},
  address={Baltimore, MD, USA},
  pages={18332--18346},
  year={2022}
}

@article{han2024fusemoe,
  title={Fusemoe: Mixture-of-experts transformers for fleximodal fusion},
  author={Han, Xing and Nguyen, Huy and Harris, Carl and Ho, Nhat and Saria, Suchi},
  journal={Adv. Neural Inf. Process. Syst.},
  volume={37},
  pages={67850--67900},
  month={Dec.},
  year={2024}
}

@article{enzweiler2011multilevel,
  title={A multilevel mixture-of-experts framework for pedestrian classification},
  author={Enzweiler, Markus and Gavrila, Dariu M},
  journal={IEEE Trans. Image Process.},
  volume={20},
  number={10},
  pages={2967--2979},
  month={Oct.},
  year={2011}
}

@article{huang2024toward,
  title={Toward efficient inference for mixture of experts},
  author={Huang, Haiyang and Ardalani, Newsha and Sun, Anna and Ke, Liu and Bhosale, Shruti and Lee, Hsien-Hsin and Wu, Carole-Jean and Lee, Benjamin},
  journal={Adv. Neural Inf. Process. Syst.},
  volume={37},
  pages={84033--84059},
  month={Dec.},
  year={2024}
}

@article{mustafa2022multimodal,
  title={Multimodal contrastive learning with limoe: the language-image mixture of experts},
  author={Mustafa, Basil and Riquelme, Carlos and Puigcerver, Joan and Jenatton, Rodolphe and Houlsby, Neil},
  journal={Adv. Neural Inf. Process. Syst.},
  volume={35},
  pages={9564--9576},
  month={Dec.},
  year={2022}
}

@article{chi2022representation,
  title={On the representation collapse of sparse mixture of experts},
  author={Chi, Zewen and Dong, Li and Huang, Shaohan and Dai, Damai and Ma, Shuming and Patra, Barun and Singhal, Saksham and Bajaj, Payal and Song, Xia and Mao, Xian-Ling and others},
  journal={Adv. Neural Inf. Process. Syst.},
  volume={35},
  pages={34600--34613},
  month={Dec.},
  year={2022}
}

@article{gale2023megablocks,
  title={Megablocks: Efficient sparse training with mixture-of-experts},
  author={Gale, Trevor and Narayanan, Deepak and Young, Cliff and Zaharia, Matei},
  journal={Proc. Mach. Learn. Syst.},
  volume={5},
  pages={288--304},
  month={Jun.},
  year={2023}
}

@article{wu2024multi,
  title={Multi-head mixture-of-experts},
  author={Wu, Xun and Huang, Shaohan and Wang, Wenhui and Ma, Shuming and Dong, Li and Wei, Furu},
  journal={Adv. Neural Inf. Process. Syst.},
  volume={37},
  pages={94073--94096},
  month={Dec.},
  year={2024}
}

@article{hwang2023tutel,
  title={Tutel: Adaptive mixture-of-experts at scale},
  author={Hwang, Changho and Cui, Wei and Xiong, Yifan and Yang, Ziyue and Liu, Ze and Hu, Han and Wang, Zilong and Salas, Rafael and Jose, Jithin and Ram, Prabhat and others},
  journal={Proc. Mach. Learn. Syst.},
  volume={5},
  pages={269--287},
  month={Jun.},
  year={2023}
}

@article{chen2022towards,
  title={Towards understanding the mixture-of-experts layer in deep learning},
  author={Chen, Zixiang and Deng, Yihe and Wu, Yue and Gu, Quanquan and Li, Yuanzhi},
  journal={Adv. Neural Inf. Process. Syst.},
  volume={35},
  pages={23049--23062},
  month={Dec.},
  year={2022}
}

@article{riquelme2021scaling,
  title={Scaling vision with sparse mixture of experts},
  author={Riquelme, Carlos and Puigcerver, Joan and Mustafa, Basil and Neumann, Maxim and Jenatton, Rodolphe and Susano Pinto, Andr{\'e} and Keysers, Daniel and Houlsby, Neil},
  journal={Adv. Neural Inf. Process. Syst.},
  volume={34},
  pages={8583--8595},
  month={Dec.},
  year={2021}
}

@article{zhou2022mixture,
  title={Mixture-of-experts with expert choice routing},
  author={Zhou, Yanqi and Lei, Tao and Liu, Hanxiao and Du, Nan and Huang, Yanping and Zhao, Vincent and Dai, Andrew M and Le, Quoc V and Laudon, James and others},
  journal={Adv. Neural Inf. Process. Syst.},
  volume={35},
  pages={7103--7114},
  month={Dec.},
  year={2022}
}

@article{han2021prediction,
  title={Prediction of combustion state through a semi-supervised learning model and flame imaging},
  author={Han, Zhezhe and Li, Jian and Zhang, Biao and Hossain, Md Moinul and Xu, Chuanlong},
  journal={Fuel},
  volume={289},
  pages={119745},
  month={Apr.},
  year={2021}
}

@article{pino2018soot,
  title={Soot propensity by image magnification and artificial intelligence},
  author={Pino, Julio and Garc{\'e}s, Hugo O and Cuevas, Juan and Arias, Luis E and Rojas, Alejandro J and Fuentes, Andr{\'e}s},
  journal={Fuel},
  volume={225},
  pages={256--265},
  month={Aug.},
  year={2018}
}

@article{zhu2019convolutional,
  title={Convolutional neural network based combustion mode classification for condition monitoring in the supersonic combustor},
  author={Zhu, Xiaobin and Cai, Zun and Wu, Jianjun and Cheng, Yuqiang and Huang, Qiang},
  journal={Acta Astronautica},
  volume={159},
  pages={349--357},
  month={Jun.},
  year={2019}
}

@inproceedings{debus2020high,
  title={High-performance data analytics of hybrid rocket fuel combustion data using different machine learning approaches},
  author={Debus, Charlotte and Ruettgers, Alexander and Petrarolo, Anna and Kobald, Mario and Siggel, Martin},
  booktitle={AIAA Scitech 2020 Forum},
  address={Orlando, FL, USA},
  pages={1161},
  year={2020}
}

@article{ren2019machine,
  title={Machine learning applied to retrieval of temperature and concentration distributions from infrared emission measurements},
  author={Ren, Tao and Modest, Michael F and Fateev, Alexander and Sutton, Gavin and Zhao, Weijie and Rusu, Florin},
  journal={Applied Energy},
  volume={252},
  pages={113448},
  month={Oct.},
  year={2019}
}

@article{dreyer2019improved,
  title={Improved methodology for performing the inverse Abel transform of flame images for color ratio pyrometry},
  author={Dreyer, Jochen AH and Slavchov, Radomir I and Rees, Eric J and Akroyd, Jethro and Salamanca, Maurin and Mosbach, Sebastian and Kraft, Markus},
  journal={Appl. Opt.},
  volume={58},
  number={10},
  pages={2662--2670},
  month={Mar.},
  year={2019}
}

@article{wei2020deep,
  title={Deep neural network inversion for 3D laser absorption imaging of methane in reacting flows},
  author={Wei, Chuyu and Schwarm, Kevin K and Pineda, Daniel I and Spearrin, R Mitchell},
  journal={Opt. Lett.},
  volume={45},
  number={8},
  pages={2447--2450},
  month={Apr.},
  year={2020}
}

@article{an2020deep,
  title={A deep learning framework for hydrogen-fueled turbulent combustion simulation},
  author={An, Jian and Wang, Hanyi and Liu, Bing and Luo, Kai Hong and Qin, Fei and He, Guo Qiang},
  journal={Int. J. Hydrogen Energy},
  volume={45},
  number={35},
  pages={17992--18000},
  month={Jul.},
  year={2020}
}

@article{broida1957spectroscopic,
  title={Spectroscopic study of electronic flame temperatures and energy distributions},
  author={Broida, Herbert P and Shuler, Kurt E},
  journal={J. Chem. Phys.},
  volume={27},
  number={4},
  pages={933--939},
  month={Oct.},
  year={1957}
}

@article{schwartz2023near,
  title={Near-MHz temperature and H2O measurements in post-detonation fireballs of 25 g hemispherical explosives using scanned-wavelength-modulation spectroscopy},
  author={Schwartz, Charles J and Stiborek, Joshua W and Butler, Austin and Chen, Damon and Guildenbecher, Daniel R and Welliver, Marc and Glumac, Nick and Goldenstein, Christopher S},
  journal={Appl. Opt.},
  volume={62},
  number={6},
  pages={1598--1609},
  month={Feb.},
  year={2023}
}

@inproceedings{sanders2008frequency,
  title={Frequency combs and hyperspectral sources for absorption spectroscopy},
  author={Sanders, Scott T},
  booktitle={Proc. Conf. Lasers Electro-Opt. (CLEO)},
  address={San Jose, CA, USA},
  pages={CMH3},
  year={2008}
}

@article{tai2021flame,
  title={Flame Temperature Measurement Based on Laser-Induced Breakdown Spectroscopy and Element Doping},
  author={Tai, Bin and Hao, Xiaojian and Wang, Jia and Sun, Haoliang},
  journal={ACS omega},
  volume={6},
  number={41},
  pages={27239--27246},
  month={Oct.},
  year={2021}
}

@article{deng2020deep,
  title={Deep learning algorithms for temperature field reconstruction of nonlinear tomographic absorption spectroscopy},
  author={Deng, Andong and Huang, Jianqing and Liu, Hecong and Cai, Weiwei},
  journal={Meas. Sens.},
  volume={10},
  pages={100024},
  month={Nov.},
  year={2020}
}

@article{kun2013development,
  title={Development of Multi-Spectral Thermometer for Explosion Flame True Temperature Measurement-Field Experiments and Measurement Accuracy Analysis},
  author={Kun, Sun and Xiao-gang, Sun and Xiao-yang, Yu and Xiao-ming, Sun},
  journal={Guangpuxue yu Guangpu Fenxi/Spectroscopy and Spectral Analysis},
  volume={33},
  pages={1719--1722},
  month={Jun.},
  year={2013}
}

@article{pei2025multispectral,
  title={Multispectral radiation thermometry based on Powell constrained optimization algorithm},
  author={Pei, Pan and Hao, Xiaojian and Feng, Shenxiang and Jia, Rui},
  journal={J. Mod. Opt.},
  volume={72},
  number={7--9},
  pages={287--294},
  month={Apr.},
  year={2025}
}

@article{wang2017tunable,
  title={Tunable diode laser absorption spectroscopy-based detection of propane for explosion early warning by using a vertical cavity surface enhanced laser source and principle component analysis approach},
  author={Wang, Yin and Wei, Yubin and Chang, Jun and Zhang, Tingting and Liu, Tongyu and Sun, Tong and Grattan, Kenneth TV},
  journal={IEEE Sensors J.},
  volume={17},
  number={15},
  pages={4975--4982},
  month={Aug.},
  year={2017}
}

@article{kranendonk2007high,
  title={High speed engine gas thermometry by Fourier-domain mode-locked laser absorption spectroscopy},
  author={Kranendonk, Laura A and An, Xinliang and Caswell, Andrew W and Herold, Randy E and Sanders, Scott T and Huber, Robert and Fujimoto, James G and Okura, Yasuhiro and Urata, Yasuhiro},
  journal={Opt. Express},
  volume={15},
  number={23},
  pages={15115--15128},
  month={Nov.},
  year={2007}
}

@article{zandieh2016sensitive,
  title={Sensitive and selective detection of adsorbed explosive molecules using opto-calorimetric infrared spectroscopy and micro-differential thermal analysis},
  author={Zandieh, Omid and Kim, Seonghwan},
  journal={Sens. Actuators B: Chem.},
  volume={231},
  pages={393--398},
  month={Aug.},
  year={2016}
}

@inproceedings{locurto2023broadband,
  title={Broadband NIR water absorption thermometry of lab scale metalized explosives},
  author={LoCurto, Anthony C and Welch, Michael and Michael, James B and Sippel, Travis R and Sims, Adam and Soo, Michael},
  booktitle={AIAA Scitech 2023 Forum},
  address={National Harbor, MD, USA},
  pages={1173},
  year={2023}
}

@article{pei2025multi,
  title={Multi-spectral imaging techniques for temperature measurement in explosion fields},
  author={Pei, Pan and Hao, Xiaojian and Wu, Ziqi and Jia, Rui and Wang, Jia and Feng, Shenxiang and Wei, Tong and You, Wenxiang and Xu, Chenyang and Wang, Xining and others},
  journal={Opt. Express},
  volume={33},
  number={8},
  pages={18180--18196},
  month={Apr.},
  year={2025}
}

@article{yu2024novel,
  title={A novel measurement strategy for explosion temperature field towards enhancing the fire process safety},
  author={Yu, Hao and Sun, Yaru and Liu, Yan and Wang, Xiaofeng and Huang, Fenglei and Liu, Haipeng},
  journal={Fire Saf. J.},
  volume={145},
  pages={104118},
  month={May.},
  year={2024}
}

@article{yao2022explosion,
  title={Explosion temperature mapping of emulsion explosives containing TiH2 powders with the two-color pyrometer technique},
  author={Yao, Yu-Le and Cheng, Yang-Fan and Zhang, Qi-Wei and Xia, Yu and Hu, Fang-Fang and Wang, Quan and Chen, Yuan},
  journal={Def. Technol.},
  volume={18},
  number={10},
  pages={1834--1841},
  month={Oct.},
  year={2022}
}

@article{hobbs2022high,
  title={High-speed infrared radiation thermometer for the investigation of early stage explosive development and fireball expansion},
  author={Hobbs, Matthew J and Barr, Andrew and Woolford, Scott and Farrimond, Dain and Clarke, Sam D and Tyas, Andrew and Willmott, Jon R},
  journal={Sensors},
  volume={22},
  number={16},
  pages={6143},
  month={Aug.},
  year={2022}
}

@inproceedings{chen2016overview,
  title={Overview of cyber-physical temperature estimation in smart buildings: From modeling to measurements},
  author={Chen, Xiaoming and Li, Xin and Tan, Sheldon X-D},
  booktitle={Proc. INFOCOM Workshops},
  address={San Francisco, CA, USA},
  pages={251--256},
  year={2016}
}

@article{nie2014temperature,
  title={Temperature measurement of gas explosion flame based on the radiation thermometry},
  author={Nie, Baisheng and He, Xueqiu and Zhang, Chen and Li, Xiangchun and Li, Hailong},
  journal={Int. J. Therm. Sci.},
  volume={78},
  pages={132--144},
  month={Apr.},
  year={2014}
}

@article{girard2014multiwavelength,
  title={Multiwavelength thermometry at high temperature: why it is advantageous to work in the ultraviolet},
  author={Girard, Ferruccio and Battuello, Mauro and Florio, M},
  journal={Int. J. Thermophys.},
  volume={35},
  number={6},
  pages={1401--1413},
  month={Jul.},
  year={2014}
}

@article{vuelban2015radiometric,
  title={Radiometric techniques for emissivity and temperature measurements for industrial applications},
  author={Vuelban, EM and Girard, Ferruccio and Battuello, Mauro and Neme{\v{c}}ek, P and Maniur, M and Pavl{\'a}sek, P and Paans, T},
  journal={Int. J. Thermophys.},
  volume={36},
  number={7},
  pages={1545--1568},
  month={Jul.},
  year={2015}
}

@article{hearst1998support,
  title={Support vector machines},
  author={Hearst, Marti A. and Dumais, Susan T and Osuna, Edgar and Platt, John and Scholkopf, Bernhard},
  journal={IEEE Intell. Syst. Their Appl.},
  volume={13},
  number={4},
  pages={18--28},
  month={Jul.},
  year={1998}
}

@article{cleveland1988locally,
  title={Locally weighted regression: an approach to regression analysis by local fitting},
  author={Cleveland, William S and Devlin, Susan J},
  journal={J. Am. Stat. Assoc.},
  volume={83},
  number={403},
  pages={596--610},
  month={Sep.},
  year={1988}
}

@article{xing2023graphical,
  title={Graphical multispectral radiation temperature inversion algorithm based on deep learning},
  author={Xing, Jian and Guo, Jiabo and Cui, Shuanglong and Li, Wenchao and Chang, Xinfang},
  journal={Opt. Lett.},
  volume={48},
  number={8},
  pages={2166--2169},
  month={Apr.},
  year={2023}
}

@article{mao2025neural,
  title={Neural-network-based multi-spectral thermometry and emissivity reconstruction in cavity high-temperature environments},
  author={Mao, Xinyu and Xie, Qi},
  journal={Appl. Opt.},
  volume={64},
  number={25},
  pages={7304--7314},
  month={Aug.},
  year={2025}
}

@inproceedings{daffara2011wide,
  title={Wide-band IR imaging in the NIR-MIR-FIR regions for in situ analysis of frescoes},
  author={Daffara, Claudia and Pezzati, L and Ambrosini, D and Paoletti, D and Di Biase, R and Mariotti, PI and Frosinini, C},
  booktitle={O3A: Optics for Arts, Architecture, and Archaeology III},
  volume={8084},
  address={Munich, Germany},
  pages={35--46},
  year={2011}
}

@article{yaseen2408yolov8,
  title={What is YOLOv8: An in-depth exploration of the internal features of the next-generation object detector},
  author={Yaseen, M},
  journal={arXiv preprint arXiv:2408.15857},
  year={2024}
}

@article{rosenblatt1958perceptron,
  title={The perceptron: a probabilistic model for information storage and organization in the brain},
  author={Rosenblatt, Frank},
  journal={Psychol. Rev.},
  volume={65},
  number={6},
  pages={386--408},
  month={Nov.},
  year={1958}
}

@article{gao2021error,
  title={Error analysis and reflection correction for radiation temperature measurements at high background temperatures},
  author={Gao, Shan and Zhao, Chunhui and Chen, Liwei and Jiang, Jing and Yu, Peifeng and Zhang, Zezhan and Wang, Chao},
  journal={Meas. Sci. Technol.},
  volume={32},
  number={5},
  pages={055003},
  month={May.},
  year={2021}
}

@article{lecun1989handwritten,
  title={Handwritten digit recognition with a back-propagation network},
  author={LeCun, Yann and Boser, Bernhard and Denker, John and Henderson, Donnie and Howard, Richard and Hubbard, Wayne and Jackel, Lawrence},
  journal={Adv. Neural Inf. Process. Syst.},
  volume={2},
  month={Nov.},
  year={1989}
}

@article{rodriguez2021retrieving,
  title={Retrieving soot volume fraction fields for laminar axisymmetric diffusion flames using convolutional neural networks},
  author={Rodr{\'\i}guez, A and Escudero, F and Cruz, JJ and Carvajal, G and Fuentes, A},
  journal={Fuel},
  volume={285},
  pages={119011},
  month={Feb.},
  year={2021}
}

@article{qi2024simulation,
	title={Simulation research of the influence of different representative wavelengths on colorimetric temperature measurement accuracy},
	author={Qi, Bowei and Jiang, Zhiwei},
	journal={J. Phys.: Conf. Ser.},
	volume={2791},
	number={1},
	pages={012016},
	year={2024}
}

@article{huang2018new,
	title={A new wavelength selection criterion for two-color pyrometer interfered with participating media},
	author={Huang, Yunwei and Long, Mujun and Chen, Dengfu and Duan, Huamei and Tan, Kai and Gui, Lintao and Liu, Tao},
	journal={Infrared Phys. Technol.},
	volume={93},
	pages={136--143},
	month={Jul.},
	year={2018}
}

@article{zhao2025,
	title={Study on a transient two-dimensional temperature measurement system for explosion fields},
	author={Zhao, Jixiang and Chen, Yang and Liu, Guanghai and Wang, Yiqing and Chang, Jing},
	journal={Optics \& Optoelectronic Technology},
	volume={23},
	number={2},
	pages={25--30},
	month={Apr.},
	year={2025},
	note={(in Chinese)}
}

@article{vallabh2022melt,
	title={Melt pool temperature measurement and monitoring during laser powder bed fusion based additive manufacturing via single-camera two-wavelength imaging pyrometry (STWIP)},
	author={Vallabh, Chaitanya Krishna Prasad and Zhao, Xiayun},
	journal={J. Manuf. Process.},
	volume={79},
	pages={486--500},
	month={Apr.},
	year={2022}
}

@article{usamentiaga2014infrared,
	title={Infrared thermography for temperature measurement and non-destructive testing},
	author={Usamentiaga, Rub{\'e}n and Venegas, Pablo and Guerediaga, Jon and Vega, Laura and Molleda, Julio and Bulnes, Francisco G},
	journal={Sensors},
	volume={14},
	number={7},
	pages={12305--12348},
	month={Jul.},
	year={2014}
}

@inproceedings{vaswani2017attention,
	title={Attention is all you need},
	author={Vaswani, Ashish and Shazeer, Noam and Parmar, Niki and Uszkoreit, Jakob and Jones, Llion and Gomez, Aidan N and Kaiser, {\L}ukasz and Polosukhin, Illia},
	booktitle={Proc. 31st Int. Conf. Neural Information Processing Systems},
	address={Long Beach, USA},
	pages={5998--6008},
	year={2017}
}

\begin{IEEEbiographynophoto}{Jie Fang}
	is currently an associate professor with the School of Communications and Information Engineering, Xi'an University of Posts and Telecommunications, Xi'an 710121, Shaanxi, P. R. China, and he is also the Chief Scientist of Hangzhou Halome Cloud Technology Company Limited, Hangzhou 310020, Zhejiang, P. R. China.
	
	His research interests include Computer Vision, Intelligent Computation, and Multi-modal Large Language model.
\end{IEEEbiographynophoto}

\begin{IEEEbiographynophoto}{Wenhao Li}
	is currently pursing his master's degree with the School of Communications and Information Engineering, Xi'an University of Posts and Telecommunications, Xi'an 710121, Shaanxi, P. R. China.
	
	His research interests include Computer Vision, Pattern Recognition, and Spectral Image Processing.
\end{IEEEbiographynophoto}

\begin{IEEEbiographynophoto}{Jianning Yang}
	is currently pursing his Ph. D with the School of Communications and Information Engineering, Xi'an University of Posts and Telecommunications, Xi'an 710121, Shaanxi, P. R. China.
	
	His research interests include Computer Vision and Few-shot Learning.
\end{IEEEbiographynophoto}

\begin{IEEEbiographynophoto}{Nan Wang}
	is currently a lecture with the School of Information Science and Technology, Hainan Normal University, Haikou 571158, Hainan, P. R. China.
	
	His research interests include Deep Learning, Computer Vision and Spectral Imagety Interpretation.
\end{IEEEbiographynophoto}

\end{document}